\PassOptionsToPackage{table}{xcolor}
\PassOptionsToPackage{pagebackref,breaklinks=true,colorlinks,citecolor=blue,urlcolor=blue,linkcolor=blue,bookmarks=false}{hyperref}
\PassOptionsToPackage{noabbrev,nameinlink,capitalize}{cleveref}

\documentclass[onecolumn, numbers]{april_aigc}

\usepackage[utf8]{inputenc} 
\usepackage{url}            
\usepackage{amsfonts}       
\usepackage{amssymb}        
\usepackage{amsmath}        
\usepackage{nicefrac}       
\usepackage{microtype}      
\usepackage{xspace}         
\usepackage{fix-cm}
\usepackage[T1]{fontenc}

\usepackage{booktabs}       
\usepackage{wrapfig}        
\usepackage{multicol}       
\usepackage{multirow}       
\usepackage{makecell}       
\usepackage{tabularx}       
\usepackage{adjustbox}      
\usepackage{colortbl}       
\usepackage{pifont}         
\usepackage{enumitem}
\usepackage[normalem]{ulem}

\newcolumntype{L}{>{\raggedright\arraybackslash}X}

\tcbuselibrary{skins, breakable}
\definecolor{blue_prompt_box}{RGB}{0, 109, 230}
\definecolor{gray_prompt_bg}{RGB}{249, 249, 249}
\newtcolorbox{promptbox}[1]{
    enhanced,
    colback=white,                   
    colframe=blue_prompt_box,
    fonttitle=\bfseries\sffamily,
    coltitle=white,                  
    title=#1,                        
    arc=3pt,                         
    outer arc=3pt,
    boxrule=0.7pt,                   
    left=6pt, right=6pt, top=6pt, bottom=6pt, 
    toptitle=1.5pt, bottomtitle=1.5pt,                
    breakable                        
}
\usepackage{xcolor}
\definecolor{linkcolor}{named}{aprilblue}
\definecolor{urlcolor}{RGB}{255,105,180}
\definecolor{citecolor}{RGB}{66,168,235}
\definecolor{lightgray}{rgb}{0.8, 0.8, 0.8}
\definecolor{darkgreen}{rgb}{0.00, 0.81, 0.78}

\definecolor{gray_tab}{RGB}{220, 220, 220}
\definecolor{blue_tab}{RGB}{227, 240, 251}
\definecolor{oran_tab}{RGB}{252, 242, 237}
\definecolor{whit_tab}{RGB}{255, 255, 255}
\definecolor{green_code}{RGB}{55, 126, 34}

\definecolor{g_scores}{RGB}{88, 142, 49}
\definecolor{r_scores}{RGB}{229, 76, 94}
\definecolor{b_scores_msfi}{RGB}{54, 108, 160}
\definecolor{p_scores_msfi}{RGB}{117, 57, 162}
\definecolor{rowgray}{rgb}{0.96, 0.96, 0.96}

\usepackage{algorithm}
\usepackage{algorithmic}
\usepackage{listings}
\usepackage{etoolbox}

\makeatletter
\AfterEndEnvironment{algorithm}{\let\@algcomment\relax}
\AtEndEnvironment{algorithm}{\kern2pt\hrule\relax\vskip3pt\@algcomment}
\let\@algcomment\relax
\newcommand\algcomment[1]{\def\@algcomment{\footnotesize#1}}
\renewcommand\fs@ruled{\def\@fs@cfont{\bfseries}\let\@fs@capt\floatc@ruled
  \def\@fs@pre{\hrule height.8pt depth0pt \kern2pt}%
  \def\@fs@post{}%
  \def\@fs@mid{\kern2pt\hrule\kern2pt}%
  \let\@fs@iftopcapt\iftrue}
\makeatother

\newcommand{\cmark}{\ding{52}\xspace}%
\newcommand{\xmarkg}{\textcolor{lightgray}{\ding{56}}\xspace}%

\def\onedot{.\xspace}
\def\eg{\textit{e.g}\onedot}

\def\etc{\textit{etc}\onedot}

\usepackage[pagebackref,breaklinks=true,colorlinks,citecolor=blue,urlcolor=blue,linkcolor=blue,bookmarks=false]{hyperref}
\AtEndPreamble{
    \usepackage[capitalize]{cleveref}
    \crefname{section}{Sec.}{Secs.}
    \Crefname{section}{Section}{Sections}
    \crefname{table}{Tab.}{Tabs.}
    \Crefname{table}{Table}{Tables}
    \crefname{equation}{Eq.}{Eqs.}
    \Crefname{equation}{Equation}{Equations}
    \crefname{figure}{Fig.}{Figs.}
    \Crefname{figure}{Figure}{Figures}
}
\hypersetup{colorlinks=true,linkcolor=linkcolor,urlcolor=urlcolor,citecolor=citecolor}

\usepackage{caption}
\DeclareCaptionFormat{custom}{{\color{aprilblue}\sffamily\textbf{#1 #2}} #3}
\titleformat*{\section}{\color{aprilblue}\Large\sffamily\bfseries}
\titleformat*{\subsection}{\color{aprilblue}\large\sffamily\bfseries}
\titleformat*{\subsubsection}{\color{aprilblue}\normalsize\sffamily\bfseries}

\usepackage{fancyhdr}
\newif\ifshowlogo
\showlogotrue   
\newcommand{\insertlogo}{%
  \ifshowlogo
    \IfFileExists{arxiv_version/assets/april_logo1.png}%
    {\includegraphics[height=0.68cm]{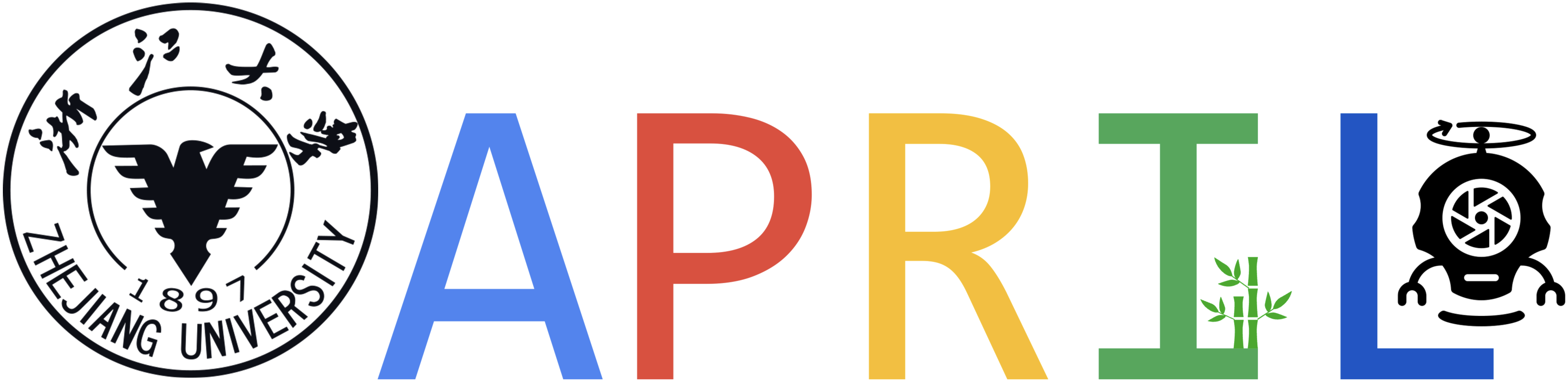}}
    {}%
  \fi
}
\newif\ifshowtoc
\showtocfalse

\def\method{PixVerve}
\def\dataset{PixVerve-95K}

\renewcommand{\title}[1]{\def\titlelist{{\fontsize{18pt}{28pt}\selectfont\sffamily\bfseries #1}}}
\title{{\method}: Advancing Native UHR Image Generation to 100MP with a Large-Scale High-Quality Dataset}

\author[1,\star]{Haojun Chen}
\author[1,\star]{Haoyang He}
\author[2,\star]{Chengming Xu}
\author[1]{Qingdong He}
\author[3]{Junwei Zhu}
\author[1]{Yabiao Wang}
\author[1]{Zhucun Xue}
\author[1]{Xianfang Zeng}
\author[3]{Zhennan Chen}
\author[4]{Xiaobin Hu}
\author[5]{Hao Zhao}
\author[1]{Yong Liu}
\author[1,\dagger,\raisebox{-0.2em}{\includegraphics[height=0.85em]{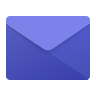}}]{Jiangning Zhang}
\author[6]{Dacheng Tao}

\affiliation[1]{Zhejiang University}
\affiliation[2]{Fudan University}
\affiliation[3]{Nanjing University}
\affiliation[4]{National University of Singapore}
\affiliation[5]{Tsinghua University}
\affiliation[6]{Nanyang Technological University}

\contribution[\star]{Joint first authors}
\contribution[\dagger]{Corresponding author}

\abstract{
Text-to-Image (T2I) models have recently seen notable progress around 1K and 2K resolution. With the extreme desire for better visual experience and the rapid development of imaging technology, the demand for Ultra-High-Resolution (UHR) image generation has grown significantly. However, UHR image generation poses great challenges due to the scarcity and complexity of high-resolution content. In this paper, we first introduce PixVerve-95K, a high-quality, open-source UHR T2I dataset curated with a carefully designed data pipeline, which contains 95K images across diverse scenarios (each image has a minimum pixel-count of 100M) and seven-dimensional annotations. Based on our large-scale image-text dataset, we take a pioneering step to extend various T2I foundation models to native 100MP generation with three training schemes. Finally, leveraging both conventional metrics and multimodal large language model-based assessments, our proposed PixVerve-Bench benchmark establishes a comprehensive evaluation protocol for UHR images encompassing visual quality and semantic alignment. Extensive experimental results on our benchmark and the constructive exploration of training strategies collaboratively provide valuable insights for future breakthroughs.
}

\coverdate{\today}
\covercorrespondence{\email{186368@zju.edu.cn}}
\coversourcecode{https://github.com/HaojunChen663/PixVerve-95K}
\coverdatasetmodelscope{https://modelscope.cn/datasets/APRIL6AIGC/PixVerve-95K}
\coverproject{https://haojunchen663.github.io/projects/PixVerve/}

\begin{document}

\maketitle
\thispagestyle{plain}

\ifshowtoc
    \clearpage
    \setcounter{tocdepth}{2} 
    
    \tableofcontents
    \vspace{1cm} 


    \clearpage
\fi

\begin{figure}[t]
    \centering
    \includegraphics[width=1\linewidth]{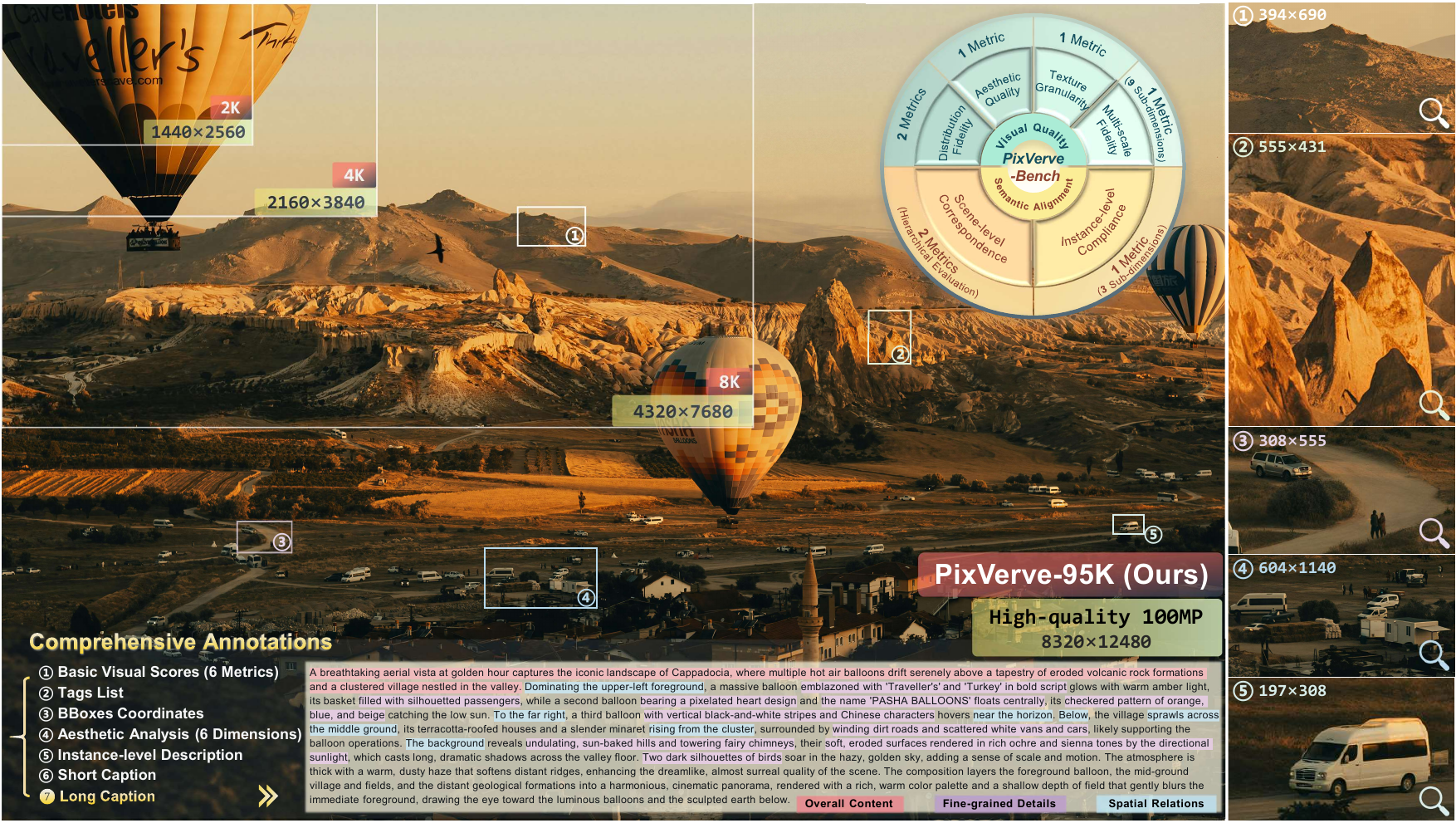}
    \caption{\textbf{PixVerve-95K} is a large-scale, high-quality dataset for Ultra-High-Resolution (UHR) image generation, first advancing Text-to-Image (T2I) generation to the 100MP scale. Featuring high visual fidelity (right) and comprehensive annotations (bottom), it can meet the growing demand for next-generation T2I applications. \textbf{PixVerve-Bench} is a comprehensive benchmark suite comprising 8 metrics for the systematic evaluation of UHR T2I methods (top-right).}
    \label{fig: teaser}
\end{figure}

\section{Introduction} \label{sec:introduction}

\begin{wrapfigure}{r}{0.465\linewidth}
    \centering
    \includegraphics[width=1\linewidth]{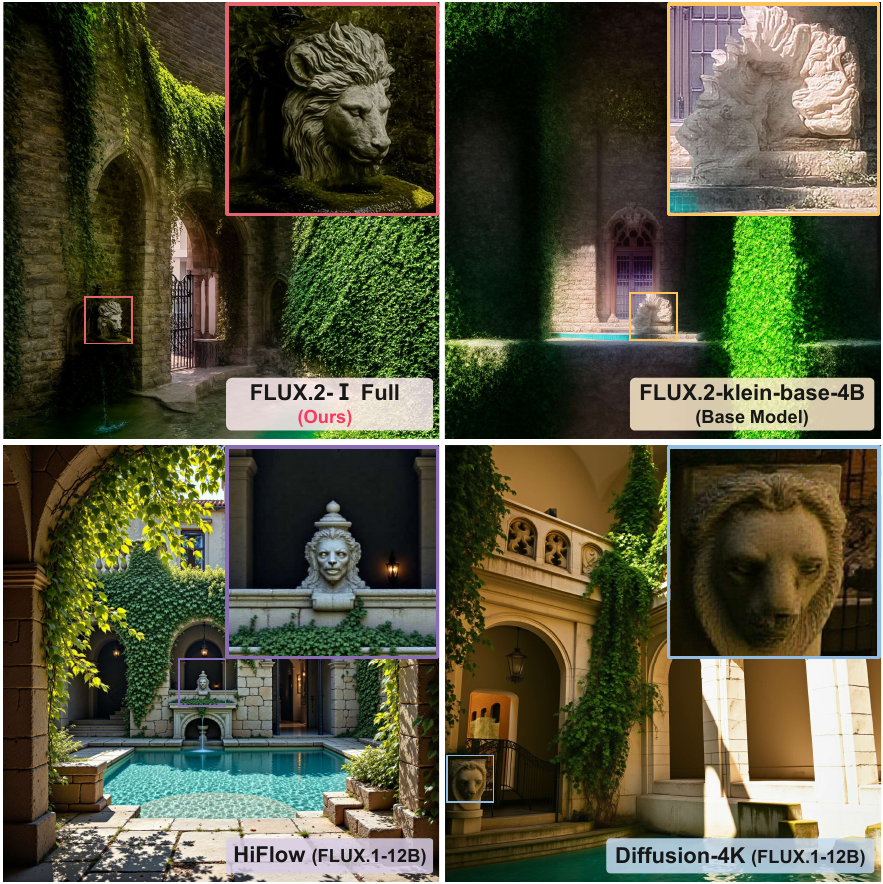}
    \caption{Visual comparison of 4K image generations. Please zoom in for clearer details.}
    \label{fig: intro_comparison}
    \vspace{-0.35em}
\end{wrapfigure}

In recent years, Text-to-Image (T2I) models have made remarkable advancements in synthesis quality and controllability~\cite{FLUX-2,Z-Image}, underscoring their exceptional potential to revolutionize the paradigm of content creation. Despite substantial progress, most existing models focus on training and generation at fixed low-to-moderate resolutions (typically 1K and 2K). Directly extrapolating these models to Ultra-High-Resolution (UHR) scenarios inevitably leads to degradations such as structural artifacts, content repetition, and a pervasive loss of high-frequency details (see \cref{fig: intro_comparison}, top-right), which significantly hinder real-world applications that necessitate photorealistic visual fidelity.

With the extreme desire for better visual experience of the next-generation media~\cite{SANA,4KAgent,UltraVideo,UltraGen,T3-Video} and empowered by computing resources, the demand for high-quality gigapixel-scale content has grown continuously in fields such as digital cinematography, immersive entertainment, and commercial design. Additionally, recent advancements in imaging technology and display devices have driven native 100-Megapixel (100MP) imaging a standard specification in modern smartphones of many brands and no longer confined to specialized domains. Furthermore, the theoretical resolution of the Human Visual System (HVS) is estimated to be 576 megapixels when integrating information across the 120-degree field of view~\cite{HumanEye}. This capacity implies that 100MP T2I generation is not merely a pursuit of larger dimensions, but a valuable quest to bridge the gap between digital synthesis and human perception. To this end, this work seeks to first advance UHR image generation to 100MP.

Recently, training-based methods for native image generation have demonstrated promising results at the 4K ($\sim$16MP) resolution~\cite{Diffusion-4K,Aesthetic-Train-V2,UltraHR-100K,UltraFlux}. Compared to training-free strategies~\cite{FouriScale,I-Max,DemoFusion,HiFlow} which often exhibit excessive smoothing and implausible details (see \cref{fig: intro_comparison}, bottom-left), these approaches enable model backbones to explicitly capture the long-range correlations within UHR images, thus attaining better performance in detail synthesis. However, extending UHR image generation to native 100MP is not simply about resolution scaling and faces three core challenges: \textbf{\textit{1)}} The primary bottleneck for native 100MP T2I training and generation lies in the lack of suitable data. Existing UHR T2I datasets are modest in resolution (typically limited to 4K~\cite{Diffusion-4K,UltraHR-100K}) due to the data scarcity and the difficulty in curating suitable data. Furthermore, public image-text corpora lack specialized captioning protocols for the UHR setting and rarely provide multi-dimensional, structured annotations which benefit precise control over various visual attributes. \textbf{\textit{2)}} The immense semantic complexity and vast pixel space of 100MP data make it challenging to design effective training schemes, which is largely unexplored in the current landscape. \textbf{\textit{3)}} Standard T2I evaluation protocols are inadequate for UHR scenarios, making it difficult to provide reliable feedback for training and model selection, as conventional metrics such as FID~\cite{FID} and CLIPScore~\cite{CLIPScore} fail to capture fine-grained details.

To bridge multi-faceted gaps, we propose a comprehensive methodology framework spanning dataset, model, and benchmark. Concretely, our core contributions are threefold:
\begin{itemize}[leftmargin=1em]
    \item We introduce \textbf{PixVerve-95K}, the first large-scale, high-quality T2I dataset to push image resolution to 100MP. With a five-stage, automated data pipeline, we curate 95,735 100MP images with fine-grained annotations (5 types of metadata and 2 comprehensive captions), directly supporting the training or fine-tuning of T2I models at high resolutions.
    \item Based on our proposed PixVerve-95K, we first explore the attempt of generating 100MP images natively. Specifically, we extend existing T2I foundation models (including both latent diffusion models and pixel diffusion models) with three distinct training schemes, providing valuable insights and paving the way for future breakthroughs.
    \item To address the limitations of conventional T2I benchmarks, we construct \textbf{PixVerve-Bench}, a systematic, hierarchical evaluation protocol incorporating both traditional metrics and assessments based on Multimodal Large Language Models (MLLMs).
\end{itemize}
\section{Related Work} \label{sec: related_work}
\subsection{Text-to-Image Datasets}
The evolution of Text-to-Image (T2I) generation has been fundamentally driven by the availability and quality of large-scale image-text datasets. The release of the early web-scale corpora such as LAION-400M~\cite{LAION-400M} and LAION-5B~\cite{LAION-5B} has significantly facilitated T2I foundation model training. As the field further matures, the focus of dataset construction starts to shift from mere volume toward high quality~\cite{Pick-a-Pic}. With the growing demand for higher resolution and visual fidelity, Diffusion-4K~\cite{Diffusion-4K} introduces the first open-source 4K T2I dataset for native UHR image training. More recently, Aesthetic-Train-V2~\cite{Aesthetic-Train-V2} and UltraHR-100K~\cite{UltraHR-100K} further expand the 4K T2I corpora. Despite these advances, most existing datasets are primarily constrained to the 1K-4K regime and often rely on global, superficial descriptions that lack the structural granularity and instance-level detail required to supervise the synthesis of exceptionally complex Ultra-High-Resolution (UHR) scenes.

\subsection{Text-to-Image Foundation Models}
Mainstream T2I foundation models include the Generative Adversarial Network (GAN)~\cite{gan}, autoregressive (AR) models~\cite{DALL-E}, and diffusion models (DMs)~\cite{DDPM}. With this evolution of architectures, DMs have recently emerged as the prevailing paradigm, pushing T2I generation to an unprecedented level~\cite{Freeu,SD,Playground-v3,FLUX-1,FLUX-2,Qwen-Image}. A pivotal milestone is the introduction of latent diffusion models (LDMs)~\cite{SD}, which perform the diffusion process in a compressed latent space, alleviating computational burdens while maintaining high perceptual fidelity~\cite{SDXL,SD3}. More recently, Diffusion Transformers (DiTs)~\cite{DiT} have made remarkable progress within the LDM framework, offering superior scalability compared to traditional U-Net backbones. Parallel to the paradigm of LDMs, pixel diffusion models perform the diffusion process directly in the raw pixel space, which have regained attention for image generation these days~\cite{JiT,DiP,PixNerd,L2P}. While LDMs are often preferred for their computational efficiency at moderate resolutions, pixel diffusion models offer a distinct advantage by bypassing the potential information loss and reconstruction artifacts inherent in Variational Autoencoder-based compression. Nevertheless, most current T2I foundation models are constrained to fixed low-to-moderate resolutions (typically 1024$\times$1024), leaving UHR T2I generation a relatively under-explored field.

\subsection{Ultra-High-Resolution Image Generation}
Beyond the 2K resolution threshold, image generation is currently dominated by LDMs. Existing solutions can be categorized into two main paradigms: training-free strategies for UHR scaling~\cite{ScaleCrafter,FouriScale,DemoFusion,HiFlow,ResMaster} and training-based methods for native UHR image generation~\cite{PixArt-sigma,UltraPixel,Diffusion-4K,Aesthetic-Train-V2,UltraHR-100K,UltraFlux,LWD}. Despite being more resource-friendly, the former approaches often suffer from object repetition, texture degradation, and unrealistic details. To enhance synthesis quality, the alternative direction curates UHR T2I corpora and trains or fine-tunes models at native high resolutions. However, current training-based frameworks remain confined to the sub-4K~\cite{PixArt-sigma} or 4K~\cite{Diffusion-4K,Aesthetic-Train-V2,UltraHR-100K,UltraFlux} scale, still falling short of the gigapixel-scale fidelity required for real-world applications. In this paper, we aim to take a pioneering step and push the frontier of T2I to the 100MP scale.
\section{Methodology: Dataset, Model, and Benchmark} \label{sec: method}
In this work, we operationalize \textit{\textbf{Native 100MP Text-to-Image Generation}} as a dedicated training and evaluation regime, significantly distinct from approaches of training-free resolution upscaling. Training-based methods treat UHR image generation as an end-to-end task that requires intrinsic high-resolution priors, while executing this regime necessitates addressing two fundamental challenges: \textit{\textbf{i)} high-quality 100MP T2I datasets and \textbf{ii)} training recipes.} Also, the absence of a systematic T2I benchmark designed for UHR scenarios hinder further research on this valuable topic. Resolving these challenges requires a holistic methodology that integrates data, model training, and evaluation.ression. 

\subsection{Curating PixVerve-95K Dataset}
\begin{figure}[t]
    \centering
    \includegraphics[width=1\linewidth]{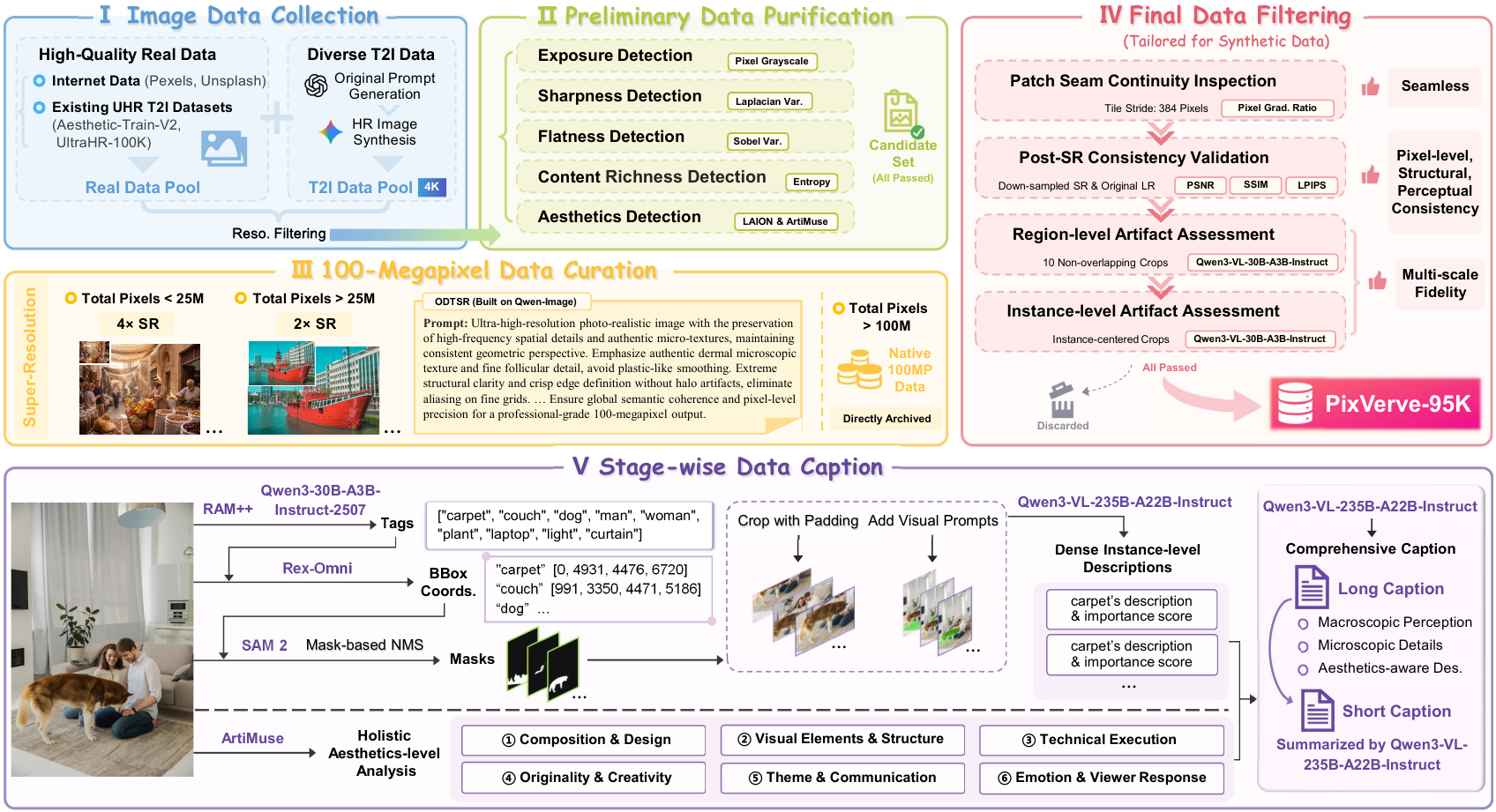}
    \caption{Overview of our \textbf{PixVerve-95K} curation pipeline that includes: \textbf{\textit{1)}} High-quality and diverse raw image data acquisition (\cref{sec: collection}). \textbf{\textit{2)}} Preliminary data purification comprising five parallel detection procedures (\cref{sec: purification}). \textbf{\textit{3)}} 100MP data curation via super-resolution (\cref{sec: SR}). \textbf{\textit{4)}} Final data filtering to ensure the quality of our synthetic data (\cref{sec: filtering}). \textbf{\textit{5)}} Stage-wise data caption pipeline carefully designed for UHR images (\cref{sec: caption}).}
    \label{fig: data_pipeline}
\end{figure}

To facilitate direct training at native 100MP resolution, we curate the first large-scale 100MP T2I dataset, addressing the critical deficit of UHR corpora in the current landscape. Beyond the pursuit of extreme resolution, we prioritize high image quality and caption comprehensiveness. To this end, we carefully design and implement a five-stage data pipeline, which is intuitively shown in \cref{fig: data_pipeline}.

\subsubsection{Raw Image Data Collection} \label{sec: collection}
\noindent \textbf{High-quality real data collection.}
To establish a large-scale image corpus for UHR T2I generation, we begin by collecting high-resolution real imagery from diverse sources. We harvest high-quality photography from platforms Pexels~\cite{pexels} and Unsplash~\cite{unsplash} via official APIs, while also integrating a subset from Aesthetic-Train-V2~\cite{Aesthetic-Train-V2} and UltraHR-100K~\cite{UltraHR-100K}. Both data collection streams are subjected to a deduplication procedure and notably, we apply the following resolution-based screening criteria to construct a data pool prior to 100MP upscaling: \textit{\textbf{i)}} total pixels exceeding 25M with a minimum dimension of 3,000 pixels, or \textit{\textbf{ii)}} total pixels ranging from 10M to 25M with a minimum dimension of 1,500 pixels. Detailed clarification on image licensing is provided in \cref{sec: appendix_license}.

\noindent \textbf{Diverse T2I data generation.}
To further enhance semantic diversity and ensure the comprehensiveness of visual concepts, we complement the real data with synthesized data. Specifically, we leverage GPT-5.1~\cite{gpt-5} to generate a set of wide-ranging, expressive prompts, which are subsequently sent to the advanced Nano Banana Pro~\cite{Gemini} to generate high-quality 4K images. Together with the real data, these diverse synthesized images constitute our raw data pool (approximately 300K).

\subsubsection{Preliminary Data Purification} \label{sec: purification}
Large-scale image corpora collected from diverse sources inevitably contain subpar samples suffering from technical degradation (\eg, exposure anomalies, blurriness, \etc), which can undermine the learning efficacy of T2I models. Therefore, to establish a baseline of visual excellence, we comprehensively evaluate each image in our raw data pool across five fundamental dimensions:

\noindent \textbf{Exposure detection.}
Overexposure and underexposure degrade the image quality greatly. Taking 5 as the threshold, we calculate the cumulative proportion of pixels with values above 250 or below 5 for each image. Any image of which the proportion exceeds 20$\%$ is deemed anomalous and excluded.

\noindent \textbf{Sharpness detection.}
To eliminate the presence of out-of-focus or motion-blurred images, we utilize the Laplacian variance as an interpretable metric for image sharpness assessment. Images yielding a score below the threshold of 10 are identified as insufficiently clear and discarded from the corpus.

\noindent \textbf{Flatness detection.}
To suppress images dominated by textureless regions, we partition each image into local patches and compute the proportion of overly smooth patches based on the Sobel variance. Images are considered to severely lack texture and then removed if the proportion exceeds 97.5$\%$.

\noindent \textbf{Content richness detection.}
Beyond basic physical properties, superior content richness is another defining characteristic of a high-quality image. We employ the classical signal, Shannon entropy~\cite{Shannon}, to quantify the informational density, retaining the top 60$\%$ highest-entropy images in the raw pool.

\begin{wrapfigure}{r}{0.58\linewidth}
    \centering
    \vspace{-0.75em}
    \includegraphics[width=1\linewidth]{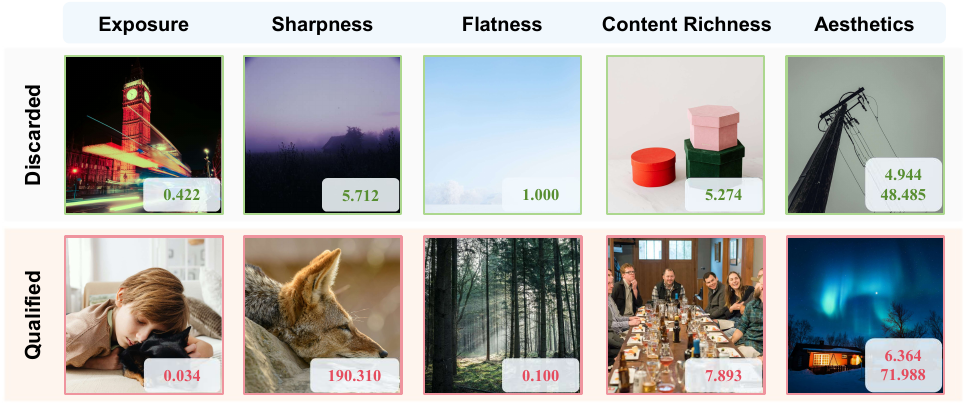}
    \vspace{-1em} 
    \caption{Representative discarded and qualified examples for each dimension. \textcolor{g_scores}{Green scores} represent disqualification, while \textcolor{r_scores}{red scores} represent passing the detection.}
    \label{fig: preliminary_data_purification}
    \vspace{-1em}
\end{wrapfigure}

\noindent \textbf{Aesthetics detection.}
Aesthetic appeal plays an important role in high-quality image generation. For aesthetics detection, we adopt a coupling approach which combines the LAION Aesthetic Predictor~\cite{LAION-Aesthetics} and ArtiMuse~\cite{ArtiMuse}, a modern MLLM-based aesthetics evaluator. We utilize both predictors to assess the aesthetic quality of each image in the raw pool with the score $S_{L}$ and $S_{A}$ respectively. Images of which $S_{L}$ or $S_{A}$ ranks among the top 60$\%$ are preserved.

By taking the intersection of the subsets retained from the five detection procedures above, the final candidate pool is derived. We present representative discarded and qualified examples for each dimension along with their corresponding scores in \cref{fig: preliminary_data_purification}, demonstrating the necessity and effectiveness of our preliminary data purification.

\subsubsection{100MP Data Curation} \label{sec: SR}
Given the scarcity of native 100MP image data in our candidate pool, we employ ODTSR~\cite{ODTSR}, a novel super-resolution (SR) framework based on Qwen-Image~\cite{Qwen-Image} which considers both fidelity and controllability, to bridge this gap and expand the scale of our final corpus. Notably, we employ a tiling strategy to tackle the high-resolution nature, which incorporates overlapping strides and feathering matrices to facilitate smooth transitions. We implement distinct upscaling intensities for different source resolutions to reach the uniform 100MP threshold, leveraging textual prompts as conditional guidance: \textbf{\textit{i)}} native 100MP images are directly archived; \textbf{\textit{ii)}} images with a total pixel count exceeding 25M are elevated via $2\times$ SR; and \textbf{\textit{iii)}} for the remaining images with total pixels in the 10M-25M range, a $4\times$ SR process is performed. This tiered production pipeline ensures that all samples achieve a minimum resolution of 100MP with high perceptual fidelity, establishing a sound data foundation for UHR T2I training and generation.

\subsubsection{Final Data Filtering} \label{sec: filtering}
To guarantee the quality of our synthetic 100MP data, we rigorously implement a four-tiered filtering pipeline, which specially targets different problems potentially introduced during the SR process.

\noindent \textbf{Patch seam continuity inspection.}
To eliminate color discontinuities and geometric misalignments, we compute the pixel gradient ratio across all horizontal and vertical seams defined by the 384-pixel tile stride used in \cref{sec: SR}. An image is considered defective and strictly excluded if any detected seam exhibits a ratio exceeding the threshold $r_{t}=2.5$.

\noindent \textbf{Post-SR consistency validation.}
To ensure pixel-level, structural, and perceptual fidelity, each synthetic 100MP image is down-sampled to its original resolution and compared against its initial input via Peak Signal-to-Noise Ratio (PSNR), Structural Similarity Index (SSIM), and LPIPS~\cite{LPIPS}. Any candidate image that fails to satisfy the tri-metric thresholds is consequently discarded.

\noindent \textbf{Region-level artifact assessment.}
To prevent local degradations such as geometric deformations and warped human features, we partition each synthetic 100MP image into non-overlapping patches of size 768 and employ a hybrid sampling strategy to select ten representative patches: six with the highest texture complexity (via the Sobel variance) and the remaining four sampled randomly. All selected patches are then evaluated by Qwen3-VL-30B-A3B-Instruct~\cite{Qwen3-VL}. An image is strictly discarded if more than one of its sampled patches is identified as containing noticeable artifacts.

\noindent \textbf{Instance-level artifact assessment.}
We further scrutinize key instances leveraging the image crops obtained in \cref{sec: caption}. Similarly, we employ Qwen3-VL-30B-A3B-Instruct~\cite{Qwen3-VL} to evaluate each crop, adopting a stringent criterion where an image is excluded if any instance is flagged as defective.

\cref{tab: dataflow} illustrates the specific data flow and the data scale at each major stage, tracing the refinement process from the initial collection to the final curated corpus.

\begin{table*}[t]
    \centering
    \caption{Data Flow and Refinement Details.}
    \label{tab: dataflow}
    \setlength\tabcolsep{17.5pt}
    \resizebox{0.92\linewidth}{!}{
    \begin{tabular}{lll}
    \toprule
    \textbf{Pipeline Stage} & \textbf{Resulting Subset} & \textbf{Scale} \\
    \midrule
    
    Image Data Collection & Raw Data Pool &  300,316 (5,000 Synthesized Images) \\
    \midrule
    
    Preliminary Data Purification & Candidate Data Pool & 122,866 \\
    \midrule
    
    Final Data Filtering & Final Data & 95,935 (95,735+200) \\
    \bottomrule
    \end{tabular}
    }
\end{table*}

\subsubsection{Stage-wise Data Caption} \label{sec: caption}
Detailed captions are crucial for fine-grained controllable image generation, which is widely recognized~\cite{DALL-E3,UltraHR-100K, UltraFlux}. However, standard zero-shot MLLM prompting often fails to encapsulate the intricate details present in UHR images. To address this challenge, we propose a hierarchical stage-wise pipeline which decouples the captioning process into three distinct layers:

\noindent \textbf{Dense instance-level descriptions generation.}
To facilitate precise alignment at the instance level, we design a cascaded pipeline utilizing the capabilities of foundation models and MLLMs. We first employ RAM$++$~\cite{RAM-plus} for open-vocabulary tagging to generate semantic tags, which are pruned by Qwen3-30B-A3B-Instruct-2507~\cite{Qwen3} to retain only tangible object tags. Rex-Omni~\cite{Rex-Omni} predicts bounding boxes (bboxes) for these filtered tags, followed by a step where SAM 2~\cite{SAM2} performs instance segmentation and generates high-fidelity masks. We further apply Non-Maximum Suppression (NMS) based on IoU to deduplicate overlapping masks and remove trivial objects with an area threshold. For context-aware captioning, we generate a visual pair for each identified instance. Specifically, we crop out a sub-image centered on the target instance with 5$\%$ padding and incorporate a highlighted prompt on the original image using its mask. These visual pairs are finally sent to Qwen3-VL-235B-A22B-Instruct~\cite{Qwen3-VL} to generate comprehensive instance-level descriptions and assign a semantic importance score to each instance.

\noindent \textbf{Holistic aesthetics-level analysis.}
Beyond instance details, a high-quality image caption should encompass an aesthetic depiction spanning multiple dimensions. To this end, we adopt ArtiMuse~\cite{ArtiMuse} to provide an expert-style aesthetic analysis across six key dimensions (composition $\&$ design, visual elements $\&$ structure, technical execution, originality $\&$ creativity, theme $\&$ communication, and emotion $\&$ viewer response), which serves as a vital reference for final caption summarization.

\noindent \textbf{Comprehensive caption summarization.}
Based on key instances' detailed descriptions and the aesthetic analysis, we employ Qwen3-VL-235B-A22B-Instruct~\cite{Qwen3-VL} as a caption synthesis expert. With the original image and all aggregated metadata, it first generates a coherent long caption encompassing the overall content and style, fine-grained details of instances, and clear relations between objects (as shown in \cref{fig: teaser}). Paired with the original image, this long caption is subsequently distilled by the same MLLM into a short caption that encapsulates the core semantic essence in a concise and fluid narrative, which can meet diverse task requirements together with the long version.

\begin{figure}[t]
    \centering
    \includegraphics[width=1\linewidth]{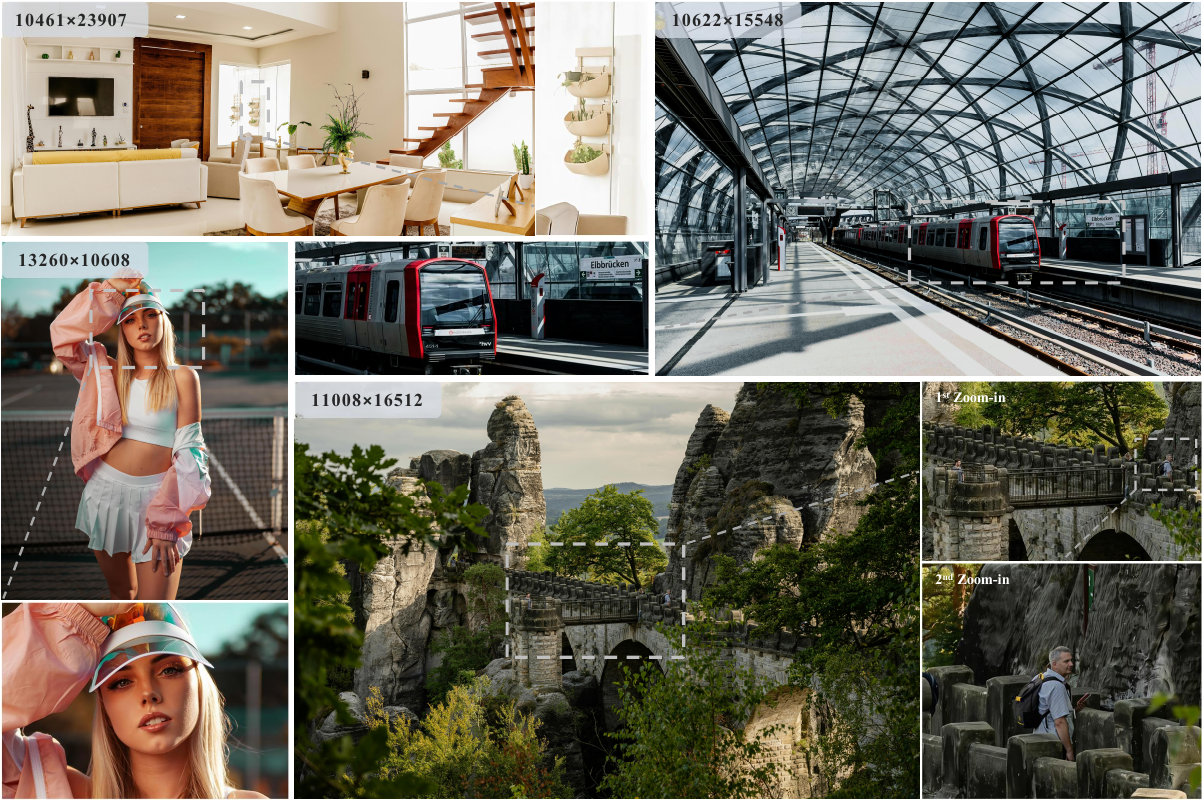}
    \caption{Qualitative samples in our PixVerve-95K dataset. The zoomed-in regions highlight the fine-grained and high-fidelity details.}
    \label{fig: dataset_samples}
\end{figure}
\begin{table*}[t]
    \centering
    \caption{Comparison with open-source UHR T2I datasets. Our proposed \textbf{\dataset} is the first premium T2I dataset to push image resolution to 10K ($\sim$100MP), providing multi-dimensional, fine-grained metadata and significantly longer comprehensive captions.}
    \label{tab: dataset}
    \renewcommand{\arraystretch}{1.175}
    \setlength\tabcolsep{3.25pt}
    \resizebox{1.0\linewidth}{!}{
    \begin{tabular}{cccccccccc}
    \hline
    \toprule[0.1em]
    Dataset & \makecell{Avg. Resolution \\ (height $\times$ width)} & Number & \makecell{Avg. Caption \\ Length} & \makecell{Basic Visual \\ Scores} & Tags & BBoxes & \makecell{Aesthetics-level \\ Analysis} & \makecell{Instance-level \\ Description} & \makecell{Variable-length \\ Caption} \\
    \hline
    
    PixArt-30k~\cite{PixArt-sigma} & 2531 $\times$ 2656 & 30,000 & 71.3 words & \xmarkg & \xmarkg & \xmarkg & \xmarkg & \xmarkg & \xmarkg   \\
    Aesthetic-Train~\cite{Diffusion-4K} & 4578 $\times$ 4838 & 12,015 & 24.2 words & \xmarkg & \xmarkg & \xmarkg & \xmarkg & \xmarkg & \xmarkg    \\
    Aesthetic-Train-V2~\cite{Aesthetic-Train-V2} & 4861 $\times$ 5127 & 105,288 & 38.1 words & \xmarkg & \xmarkg & \xmarkg & \xmarkg & \xmarkg & \xmarkg    \\
    UltraHR-100K~\cite{UltraHR-100K} & 3654 $\times$ 5143 & 100,486 & 109.2 words & \xmarkg & \xmarkg & \xmarkg & \xmarkg & \xmarkg & \xmarkg    \\
    \hline

    \rowcolor{blue_tab}
    \textbf{PixVerve-95K (Ours)} & \textbf{13031 $\times$ 15348} & 95,735 & \begin{tabular}[c]{@{}c@{}}234.1 words \\ (Long)\end{tabular} & \begin{tabular}[c]{@{}c@{}}\cmark \\ (6 Metrics)\end{tabular} & \cmark & \cmark & \begin{tabular}[c]{@{}c@{}}\cmark \\ (6 Dimensions)\end{tabular} & \cmark & \cmark   \\
    \hline
    \toprule[0.1em]
    \end{tabular}
    \vspace{-3.05em}
}
\end{table*}
\begin{figure}[t]
    \includegraphics[width=1\linewidth]{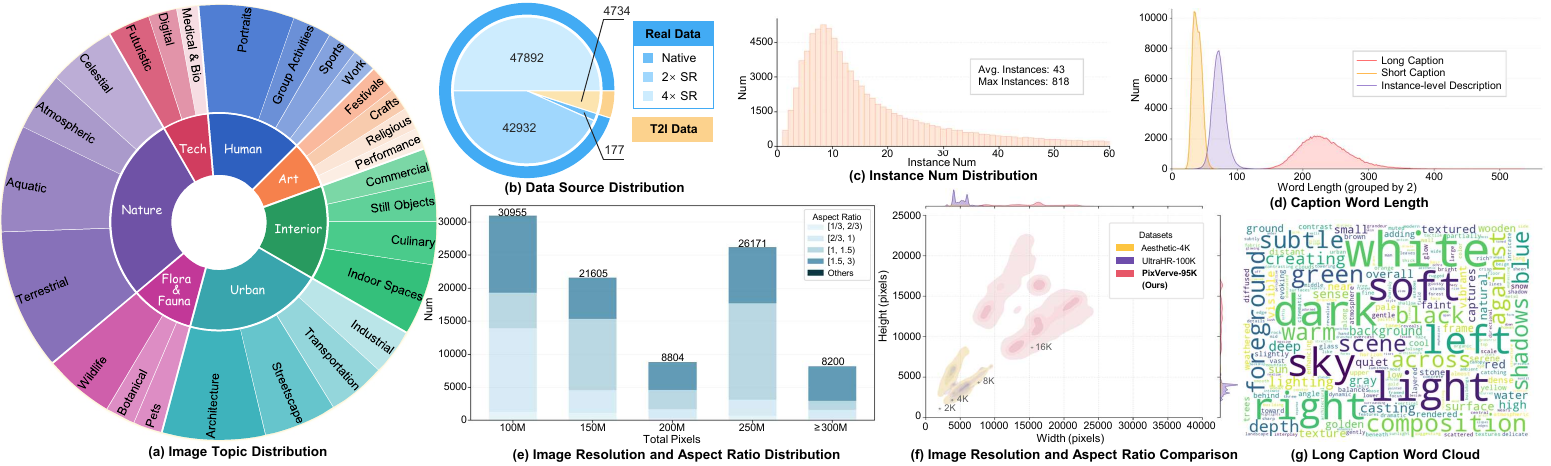}
    \caption{Statistical distributions of our PixVerve-95K.}
    \label{fig: statistical_distributions}
\end{figure}

\subsubsection{Statistical Comparison and Analysis}
With our five-stage data pipeline, we construct \textbf{PixVerve-95K}, comprising 95,735 100MP images with comprehensive annotations. \cref{fig: dataset_samples} presents some qualitative samples, which are best viewed zoomed-in. As summarized in \cref{tab: dataset}, PixVerve-95K is the first to push open-source T2I data to 10K resolution ($\sim$100MP), providing five-dimensional metadata (basic visual scores, tags, bboxes, aesthetics-level analysis, and instance-level description) beyond long and short captions. These structured annotations offer versatile utility for the community, enabling granular control over data quality and facilitating adaptive sampling strategies tailored to specialized training objectives. \cref{fig: statistical_distributions} visualizes the statistical distributions of PixVerve-95K from multiple perspectives, highlighting the scenario diversity, balanced aspect ratios, and the exceptional expressiveness of aggregated captions. 
\subsection{Extending Text-to-Image Foundation Models to Native 100MP Generation} \label{sec: model}
Training T2I foundation models at native 100MP poses great challenges due to the immense semantic complexity and vast pixel space. Bridging this gap necessitates a collaborative exploration of diverse architectural and optimization strategies rather than a singular fix. Based on PixVerve-95K, we conduct a multi-faceted exploration to retrofit T2I foundation models for native 100MP synthesis. Specifically, we investigate three distinct schemes to identify the optimal path:

\begin{itemize}[leftmargin=1em]
    \item \textbf{\textit{Scheme $\mathrm{I}$}: Full-Attention LDM Fine-tuning.} As direct baselines, we perform full-parameter training and use LoRA for parameter-efficient fine-tuning (PEFT) on FLUX.2-klein-base-4B~\cite{FLUX-2}, respectively. While maximizing the retention of the pre-trained model's semantic prior and structural integrity, this approach encounters severe hardware constraints. The latent representation dimensions lead to an exponential surge in memory requirements, mandating model parallelism for inference at the 100MP scale, which limits the flexibility for general-purpose deployment in real world.
    \item \textbf{\textit{Scheme $\mathrm{II}$}: Window-Attention Retrofitting and LDM Fine-tuning.} Inspired by EMOv2~\cite{EMOv2}, we refine the training strategy by introducing a local-to-global attention mechanism. We retrofit the joint attention in FLUX.2 into a dual-branch window-attention, without altering the core architecture of full-attention pretrained models. Specifically, text queries attend to all text and image tokens, while each image query attends to all text tokens and two complementary image-token neighborhoods. The close branch partitions the latent grid into contiguous spatial windows, preserving high-frequency local structure; while the remote branch groups tokens with the same modulo offset under the same partition stride, providing sparse long-range communication across the canvas. The outputs of the close and remote branches are averaged to approximate the original full image attention. For a latent grid consisting of $N=HW$ image tokens, a standard full self-attention mechanism incurs a quadratic computational cost of $\mathcal{O}(N^{2})$. By retrofitting the attention with partition factors $(a,b)$, this scheme effectively reduces the image-image attention complexity to approximately $\mathcal{O}(2N^2/(ab))$, while the complexity for text-image conditioning remains linear with respect to $N$ since the number of text tokens is small. Following T3-Video~\cite{T3-Video}, we further cycle layer-wise partition schedules with different window size so different blocks exchange information under different receptive-field shapes.
    \item \textbf{\textit{Scheme $\mathrm{III}$}: Patch-based Diffusion in Pixel Space.} Motivated by recent pixel diffusion models~\cite{JiT,DiP,L2P}, we explore a paradigm bypassing the latent space entirely. Following L2P~\cite{L2P}, which is built on DiP~\cite{DiP}, we adopt a patch-based pixel diffusion framework that decouples global structure from local refinement: a transformer backbone operates on large image patches for long-range semantics and spatial layout, while a lightweight head leverages contextual features and original noisy patches to reconstruct fine details. From the theoretical perspective of L2P, large-patch tokenization preserves global low-frequency information efficiently, but high-frequency components are only weakly recovered during denoising unless explicit local inductive bias is introduced, making dedicated patch refinement crucial for faithful pixel-space reconstruction. However, scaling this scheme to UHR scales exposes a severe sequence-length and memory bottleneck. To enable training on a single 96 GB GPU card, we adaptively adjust the patch size to control the token count at the cost of coarser patch-level representations at higher resolutions.
\end{itemize}

\noindent \textbf{Progressive Training Strategy.} To mitigate the training instability inherent in the transition from standard resolutions to 100MP, we implement a three-stage progressive training strategy across all schemes. Models are fine-tuned through three graduated resolution tiers: 4K ($\sim$16MP), 8K ($\sim$64MP), and finally the target 10K (100MP). Concretely, these constructive experimental routes collaboratively provide critical insights into the resolution scalability of current T2I foundation models.
\subsection{PixVerve-Bench Construction and Evaluation}
For systematic and universal evaluation of UHR T2I models, we introduce \textbf{PixVerve-Bench}, comprising 200 manually picked images across diverse scenarios with an average resolution of 12369$\times$14377. The benchmark framework (see \cref{fig: teaser}, top-right) leverages both conventional metrics and novel MLLM-as-a-judge protocols, providing a holistic evaluation across two complementary aspects: \textbf{\textit{1)} Visual Quality Assessment}, comprising four critical dimensions, including distribution consistency, aesthetic quality, texture granularity, and multi-scale fidelity; and \textbf{\textit{2)} Semantic Alignment Evaluation}, which assesses instructional adherence across scene-level correspondence and instance-centric compliance. Detailed evaluation procedures and scoring formulas are provided in \cref{sec: appendix_metric}.

\subsubsection{Visual Quality Assessment}
Visual Quality evaluates the intrinsic physical attributes and perceptual realism of the generated images across the following four dimensions:

\noindent \textbf{Distribution Consistency.} Following common practice, we employ FID~\cite{FID} to evaluate the overall distribution fidelity of the generated images. Considering that FID is calculated on down-sampled (299$\times$299) images and neglects details, we incorporate $\text{FID}_\text{patch}$ to scrutinize local patches.

\noindent \textbf{Aesthetic Quality.} Utilizing the LAION Aesthetic Predictor~\cite{LAION-Aesthetics}, we map each generated image into an aesthetic feature space and score its aesthetic quality.

\noindent \textbf{Texture Granularity} focuses on the richness and complexity of micro-patterns, which are among the most significant characteristics of UHR images. Using the Gray Level Co-occurrence Matrix (GLCM)~\cite{GLCM} Score, we provide a rigorous diagnosis of whether the generated images suffer from monotonous flatness. Higher scores indicate richer texture and greater details.

\noindent \textbf{Multi-scale Fidelity.} The fidelity of UHR images is susceptible to both global artifacts (\eg, structural incoherence and physical distortion) and local artifacts (\eg, unrealistic noise and pattern repetition), which are difficult to capture using conventional metrics. Therefore, we employ Qwen3.5-35B-A3B~\cite{Qwen3.5} to perform a rigorous multi-scale fidelity assessment. Specifically, we systematically categorize the UHR-specific artifacts into two main dimensions and nine fine-grained sub-dimensions. The model is instructed to assign a score on a five-point scale for each sub-dimension based on the severity of the artifact, followed by a unifying step where these individual scores are integrated into an interpretable metric, the \textbf{Multi-scale Fidelity Index (MSFI)}, to reflect overall performance. 
   
\subsubsection{Semantic Alignment Evaluation}
Semantic Alignment evaluates how well the generated visual content adheres to the provided textual prompts. Given the expansive canvas and semantic complexity in UHR T2I generation, we hierarchically assess instructional adherence across two levels of granularity:

\noindent \textbf{Scene-level Correspondence.} For this foundational granularity, we first utilize CLIPScore~\cite{CLIPScore} to measure the global semantic correlation using the short captions. Furthermore, we incorporate FG-CLIP2 Score~\cite{FG-CLIP2} to better capture fine-grained details, which is computed on the long captions.

\noindent \textbf{Instance-centric Compliance.} To complement global correlation metrics, we propose the \textbf{Instance-centric Compliance Score (ICS)}. ICS leverages the advanced capabilities of Qwen3.5-35B-A3B~\cite{Qwen3.5} to assess semantic alignment across three hierarchical dimensions: Instance Existence Verification, Appearance Attribute Alignment, and Spatial Relation Accuracy, which provides a fine-grained and interpretable metric for measuring whether visual elements adhere to textual prompts.
\section{Experiments} \label{sec: experiment}
\subsection{Experimental Setup}
\noindent \textbf{Overall training settings.} As introduced in \cref{sec: model}, we investigate three training schemes in our experiments. \textbf{\textit{Scheme $\mathrm{I}$}:} We fine-tune the pretrained FLUX.2-klein-base-4B~\cite{FLUX-2} model using both full-parameter tuning and LoRA adaptation. For full-parameter fine-tuning, the learning rate is fixed at $1 \times 10^{-5}$ across all resolution tiers. For LoRA fine-tuning, we use a learning rate of $1 \times 10^{-4}$ and set the LoRA rank to 32. \textbf{\textit{Scheme $\mathrm{II}$}:} For the window-attention retrofitting, we adopt window aspect ratios of $1\!:\!1$, $1\!:\!2$, $2\!:\!1$, $1\!:\!8$, and $8\!:\!1$. The window size is scaled linearly with the input resolution. \textbf{\textit{Scheme $\mathrm{III}$}:} To enable training on a single 96 GB GPU card at different scales, we adjust the patch size used in L2P~\cite{L2P} accordingly. Specifically, the patch sizes are set to 64, 128, and 320 for 4K, 8K, and 10K resolution, respectively. Unless otherwise specified, the learning rate is uniformly set to $5 \times 10^{-5}$. More training details including the fine-tuning epochs and computational expenditures are provided in \cref{sec: appendix_train}.

\noindent \textbf{Baselines.}
Corresponding with our proposed training schemes, we denote our fine-tuned variants as FLUX.2-$\mathrm{I}$ (Full), FLUX.2-$\mathrm{I}$ (LoRA), FLUX.2-$\mathrm{II}$, and L2P-$\mathrm{III}$, respectively. To comprehensively evaluate our approaches, we conduct extensive comparisons against different baselines across three UHR scales: 4K (4096$\times$4096), 8K (8192$\times$8192), and 10K (10240$\times$10240). The compared methods encompass: \textbf{\textit{i)}} direct extrapolation of pre-trained T2I models (FLUX.2-klein-base-4B~\cite{FLUX-2}, Qwen-Image~\cite{Qwen-Image}, and L2P~\cite{L2P}), \textbf{\textit{ii)}} representative training-free strategies (DemoFusion~\cite{DemoFusion}, LinFusion~\cite{LinFusion}, and HiFlow~\cite{HiFlow}), and \textbf{\textit{iii)}} recent training-based models (UltraPixel~\cite{UltraPixel}, UltraFlux~\cite{UltraFlux}, and Diffusion-4K~\cite{Diffusion-4K}). All baselines are evaluated with their official implementations and parameter settings.

\subsection{Experimental Results, Observations, and Analysis}
\begin{figure}[t]
    \centering
    \includegraphics[width=1\linewidth]{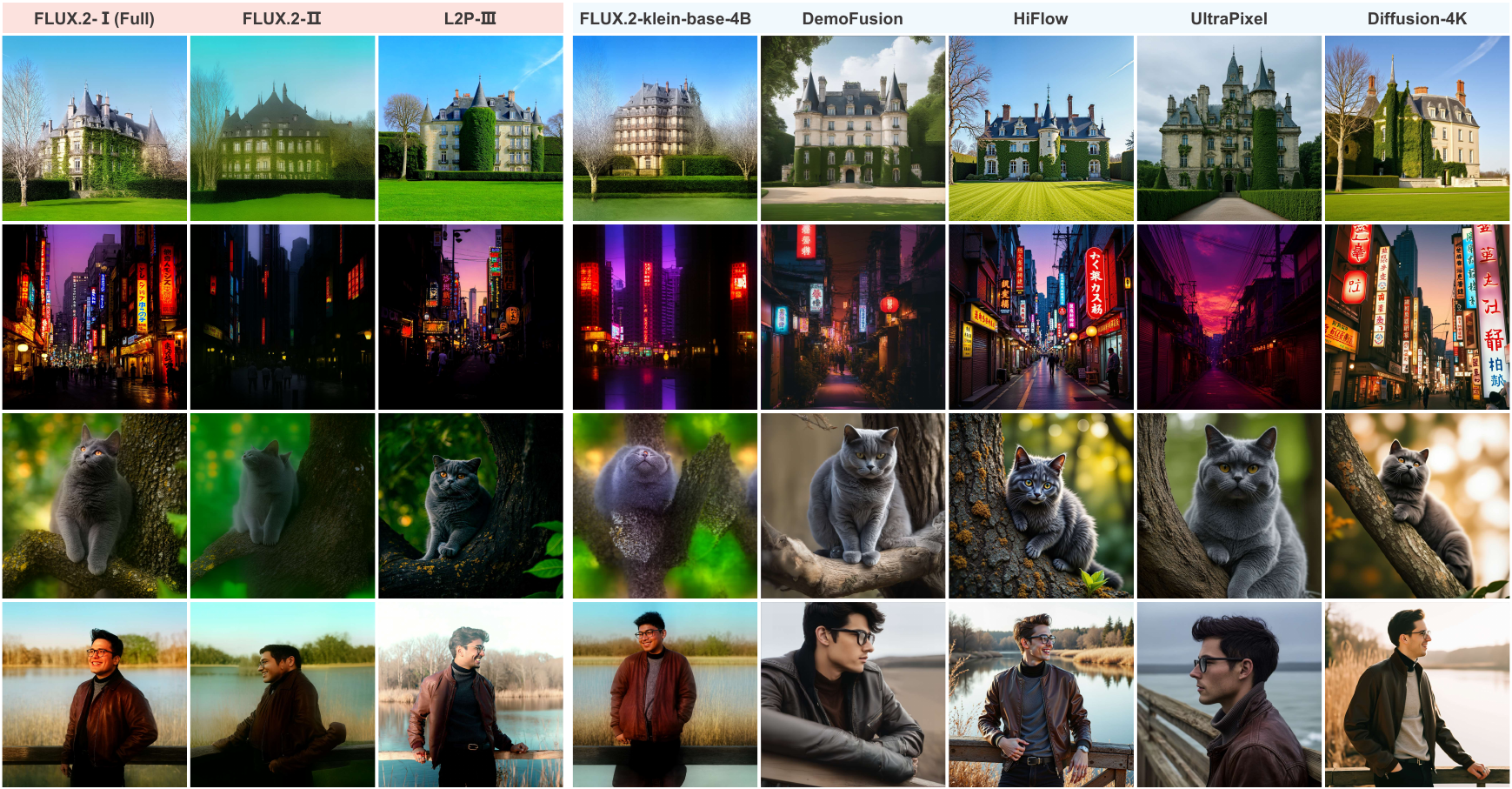}
    \caption{Qualitative comparison of different methods at 4K (4096$\times$4096) resolution.}
    \label{fig: 4K_comparison}
\end{figure}

\cref{tab: benchmark_results} presents the quantitative performance across three different resolutions on PixVerve-Bench, with detailed sub-dimension performance of the MSFI shown in \cref{tab: detailed_performance}). \cref{fig: 4K_comparison} as well as \cref{fig: intro_comparison} illustrates a qualitative comparison at 4K resolution. In this section, we report our key observations and analysis regarding the resolution scalability of current methods and different T2I foundation models.

\begin{table*}[t]
    \centering
    \caption{\textbf{Quantitative comparison on PixVerve-Bench.} The best result is highlighted in \textbf{bold}, while the second-best result is \underline{underlined}. -- indicates complete failures such as producing meaningless textures or black images, which are not applicable to the semantics-agnostic GLCM Score and MSFI.}
    \label{tab: benchmark_results}
    \renewcommand{\arraystretch}{1.15}
    \setlength\tabcolsep{4.25pt}
    \resizebox{1.0\linewidth}{!}{
    \begin{tabular}{cc|ccccc|ccc}
    \hline
    \toprule[0.1em]
    \multirow[c]{3}{*}{\makecell{\textbf{Resolution} \\ \textbf{(height $\times$ width)}}} & 
    \multicolumn{1}{c|}{\multirow[c]{3}{*}{\textbf{Method}}} & 
    \multicolumn{5}{c|}{\textbf{Visual Quality}} & 
    \multicolumn{3}{c}{\textbf{Semantic Alignment}} \\ 
    \cmidrule(l){3-10}
    \multicolumn{2}{c|}{} & FID $\downarrow$ & $\text{FID}_\text{patch} \downarrow$ & Aesthetics $\uparrow$ & \makecell{GLCM \\ Score} $\uparrow$ & MSFI $\uparrow$ & CLIPScore $\uparrow$ & \makecell{FG-CLIP2 \\ Score} $\uparrow$ & ICS $\uparrow$\\
    \hline
    
    \multirow[c]{13}{*}{\makecell{\textbf{4K} \\ \textbf{(4096 $\times$ 4096)}}} & 
    FLUX.2-klein-base-4B~\cite{FLUX-2} & 167.234 & 76.794 & 5.498 & 0.873 & 7.408 & 31.058 & 17.049 & 5.376 \\
    & Qwen-Image~\cite{Qwen-Image} & 140.740 & 53.200 & 5.707 & 0.611 & 8.293 & 33.041 & 18.492 & 6.753 \\
    & L2P~\cite{L2P} & 126.089 & 90.803 & 5.852 & 0.818 & 6.904 & 33.890 & 19.501 & 7.954 \\
    & DemoFusion~\cite{DemoFusion} & 142.981 & 55.164 & 6.167 & 0.733 & 8.246 & 32.608 & 18.105 & 3.619 \\
    & LinFusion~\cite{LinFusion} & 142.933 & 68.184 & \textbf{6.302} & 0.394 & 8.171 & 32.246 & 17.694 & 3.849 \\
    & HiFlow~\cite{HiFlow} & 130.337 & 49.842 & 6.189 & \textbf{1.068} & 8.779 & 34.190 & 19.643 & 6.978 \\
    & UltraPixel~\cite{UltraPixel} & 144.859 & 66.878 & \underline{6.260} & 0.732 & \textbf{9.153} & 32.430 & 17.836 & 3.741 \\
    & UltraFlux~\cite{UltraFlux} & \underline{121.337} & 49.902 & 6.068 & \underline{1.037} & 8.712 & \textbf{34.909} & \textbf{20.084} & \underline{8.530} \\
    & Diffusion-4K~\cite{Diffusion-4K} & 134.702 & 78.323 & 5.848 & 0.668 & 8.377 & 33.421 & 18.749 & 6.423 \\
    \rowcolor{blue_tab} \cellcolor{white} & FLUX.2-$\mathrm{I}$ (Full) & 128.897 & \underline{45.204} & 5.804 & 0.987 & 8.911 & 34.161 & 19.683 & \textbf{8.533} \\
    \rowcolor{blue_tab} \cellcolor{white} & FLUX.2-$\mathrm{I}$ (LoRA) & 127.436 & \textbf{40.433} & 5.798 & 0.977 & \underline{8.977} & 34.178 & \underline{19.767} & 8.420 \\
    \rowcolor{blue_tab} \cellcolor{white} & FLUX.2-$\mathrm{II}$ & 161.729 & 76.460 & 5.530 & 0.593 & 8.125 & 31.663 & 18.119 & 5.340 \\
    \rowcolor{blue_tab} \cellcolor{white} & L2P-$\mathrm{III}$ & \textbf{118.183} & 98.704 & 5.792 & 0.896 & 6.990 & \underline{34.264} & 19.676 & 7.870 \\
    \hline
    \multirow[c]{6.5}{*}{\makecell{\textbf{8K} \\ \textbf{(8192 $\times$ 8192)}}} & 
    FLUX.2-klein-base-4B~\cite{FLUX-2} & 422.737 & 350.331 & 4.121 & -- & -- & 18.345 & 0.503 & 0.318 \\ 
    & L2P~\cite{L2P} & \underline{140.396} & 87.167 & 5.590 & 0.363 & \underline{7.592} & \textbf{33.261} & \textbf{18.548} & \textbf{6.394} \\
    & DemoFusion~\cite{DemoFusion} & 176.068 & \textbf{58.480} & \underline{6.031} & \underline{0.514} & 6.966 & 31.529 & 17.057 & 2.122 \\
    & LinFusion~\cite{LinFusion} & 143.429 & \underline{62.658} & \textbf{6.271} & 0.177 & \textbf{7.797} & 32.097 & 17.530 & 3.809 \\
    \rowcolor{blue_tab} \cellcolor{white} & FLUX.2-$\mathrm{I}$ (Full) & 197.690 & 66.571 & 5.289 & -- & -- & 28.676 & 10.765 & 0.420 \\ 
    \rowcolor{blue_tab} \cellcolor{white} & FLUX.2-$\mathrm{I}$ (LoRA) & 277.410 & 99.414 & 4.752 & -- & -- & 24.250 & 8.086 & 0.411 \\ 
    \rowcolor{blue_tab} \cellcolor{white} & L2P-$\mathrm{III}$ & \textbf{134.635} & 133.453 & 5.569 & \textbf{1.122} & 5.310 & \underline{32.788} & \underline{17.802} & \underline{5.504} \\
    \hline
    \multirow[c]{4}{*}{\makecell{\textbf{10K} \\ \textbf{(10240 $\times$ 10240)}}} & 
    L2P~\cite{L2P} & \underline{156.158} & 116.379 & 5.438 & 0.533 & \underline{7.397} & \textbf{32.440} & \textbf{17.866} & \textbf{5.548} \\
    & DemoFusion~\cite{DemoFusion} & 179.063 & \textbf{61.854} & \underline{5.895} & \underline{0.538} & 6.710 & 30.930 & 15.920 & 2.671 \\
    & LinFusion~\cite{LinFusion} & \textbf{152.964} & \underline{70.718} & \textbf{6.215} & 0.156 & \textbf{7.525} & \underline{31.937} & \underline{17.380} & \underline{4.825} \\
    \rowcolor{blue_tab} \cellcolor{white} & L2P-$\mathrm{III}$ & 159.212 & 192.286 & 5.569 & \textbf{1.100} & 5.567 & 31.810 & 17.057 & 3.586 \\

    \hline
    \toprule[0.1em]
    \end{tabular}
    }
\end{table*}

\noindent\textbf{Base Model and Existing Methods.}
Directly extrapolating the base model FLUX.2-klein-base-4B to UHR generation leads to severe degradation: at 8K, FID exceeds 422, and CLIPScore drops to 18.345. Training-free strategies such as DemoFusion~\cite{DemoFusion} and LinFusion~\cite{LinFusion} can maintain relatively good local visual statistics beyond 4K, \eg, DemoFusion obtains the best $\mathrm{FID}_{\mathrm{patch}}$ at 8K and 10K. However, their performance on semantic alignment remains limited; DemoFusion achieves ICS below 3.7 across all resolutions, indicating that tiled or progressive inference struggles to preserve prompt consistency. At 4K, UltraFlux~\cite{UltraFlux} is a strong baseline with FID 121.337 and ICS 8.530, but our FLUX.2-$\mathrm{I}$ achieves better local fidelity with lower $\mathrm{FID}_{\mathrm{patch}}$.

\noindent\textbf{\textit{Scheme} $\mathrm{I}$: Strong 4K Adaptation but Poor Scalability.}
Fine-tuning the base model with \textit{Scheme} $\mathrm{I}$ achieves the best balance at 4K. Compared with the base FLUX.2-klein-base-4B, the LoRA variant reduces $\mathrm{FID}_{\mathrm{patch}}$ from 76.794 to 40.433, while the full-parameter variant improves ICS from 5.376 to 8.533. These results show that fine-tuning effectively adapts local image statistics while preserving the semantic prior. Qualitatively, it generates coherent layouts and fine details without over-sharpening or texture collapse. However, this advantage comes with heavy computational overheads. As shown in \cref{tab: appendix_training_details}, full fine-tuning at 4K costs over 20,000 H20 GPU hours, and only 0.25 epoch at 8K costs over 10,000 GPU hours. Inference also scales poorly: \cref{tab: appendix_inference_details} shows that FLUX.2-$\mathrm{I}$ variants require 8 GPUs and over 100s at 4K, over 1,200s at 8K, and nearly 3,000s at 10K. This confirms that full-attention LDM fine-tuning is effective at 4K but impractical for native 100MP synthesis.

\noindent\textbf{\textit{Scheme} $\mathrm{II}$: Faster Attention with Optimization Difficulty.} FLUX.2-$\mathrm{II}$ reduces computational cost through window-attention retrofitting. As shown in \cref{tab: appendix_inference_details}, the inference time at 4K is reduced by approximately 32s on 8 GPUs, giving 30\% speedup. Its 4K training cost is also lower than that with \textit{Scheme} $\mathrm{I}$, requiring 9,216 H20 GPU hours for 3 epochs. However, we observe that performance drops noticeably: FLUX.2-$\mathrm{II}$ obtains $\mathrm{FID}_{\mathrm{patch}}$ 76.460 and ICS 5.340, close to the base model but much worse than both FLUX.2-$\mathrm{I}$ variants. This suggests a mismatch between the pretrained full-attention structure and the retrofitted local attention pattern. In practice, the model requires more optimization steps to recover global communication, and the final quality is sensitive to the window size and aspect-ratio schedules.

\noindent\textbf{\textit{Scheme} $\mathrm{III}$: Best Scalability with a Patch-Size Trade-off.} It is surprisingly observed that L2P-$\mathrm{III}$ shows the strongest scalability among our explored schemes. It achieves the best FID at 4K and 8K, with scores of 118.183 and 134.635, and remains functional at 10K. More importantly, \cref{tab: appendix_inference_details} shows that L2P-$\mathrm{III}$ needs one GPU and 58s, 70s, and 88s for 4K, 8K, and 10K inference. Compared with DemoFusion, it is up to $156\times$ faster. Compared with FLUX.2-$\mathrm{I}$ LoRA at 10K, it is over $33\times$ faster while using 1 GPU instead of 8. The main limitation is the patch-size trade-off: to fit training into a single 96 GB GPU card, the patch size increases from 64 to 128 and 320, reducing memory cost but weakening fine-detail reconstruction. This explains its less competitive $\mathrm{FID}_{\mathrm{patch}}$ and MSFI at high resolutions. Overall, patch-based pixel diffusion is currently the most practical path toward native 100MP generation.

\subsection{Ablation Study and Discussion}
\noindent \textbf{Ablation on image caption quality.}
We investigate the impact of caption granularity on image synthesis by comparing 4K image generations under short and long prompting settings on PixVerve-Bench. Evaluations across UltraFlux, Diffusion-4K, and FLUX.2-$\mathrm{I}$ (Full) reveal consistent performance gains when using long captions, as summarized in \cref{tab: ablation_captions}, indicating that increased descriptive granularity and semantic density are of great significance for UHR image generation. The visual comparison is shown in \cref{fig: appendix_ablation_captions}.

\begin{wraptable}{r}{0.5\textwidth}
    \centering
    \vspace{-0.95em}
    \caption{Ablation on image caption quality.}
    \label{tab: ablation_captions}
    \renewcommand{\arraystretch}{1.15}
    \setlength\tabcolsep{2pt}
    \resizebox{1.0\linewidth}{!}{
        \begin{tabular}{cc|cccc}
            \toprule[0.15em]
            Method & Caption & FID $\downarrow$ & $\text{FID}_\text{patch} \downarrow$ & Aesthetics $\uparrow$ & GLCM Score $\uparrow$ \\
            \midrule
            \multirow{2}{*}{UltraFlux} & Short & 126.316 & 55.732 & 6.058 & 1.008 \\
            & Long & \textbf{121.337} & \textbf{49.902} & \textbf{6.068} & \textbf{1.037} \\
            \midrule
            \multirow{2}{*}{Diffusion-4K} & Short & 142.628 & 90.728 & 5.748 & 0.606 \\
            & Long & \textbf{134.702} & \textbf{78.323} & \textbf{5.848} & \textbf{0.668} \\
            \midrule
            \multirow{2}{*}{FLUX.2-$\mathrm{I}$ (Full)} & Short & 135.837 & 51.173 & 5.793 & 0.893 \\
            & Long & \textbf{128.897} & \textbf{45.204} & \textbf{5.804} & \textbf{0.987} \\
            \toprule[0.15em]
        \end{tabular}
    }
    \vspace{-1em}
\end{wraptable}

\noindent \textbf{Discussion on downstream applications.}
Beyond T2I generation, our dataset supports diverse downstream tasks such as UHR image quality assessment, fine-grained understanding in UHR contexts, image outpainting, and benchmarking for image compression. To validate one of these potential utilities, we conduct a performance evaluation of different lossless image compressed formats using our 100MP images. Compared to small images for which program startup time accounts significantly during testing, time consumption on large images can better reflect the true performance of the codecs. Detailed experimental setup and results are provided in \cref{sec: appendix_compression}.

\section{Conclusion} 
\label{sec:conclusion}
In this paper, we present a comprehensive framework to explore the frontier of native 100MP T2I generation, encompassing data, training, and evaluation. To address key challenges in data scarcity and semantic granularity, we introduce PixVerve-95K, a high-quality, open-source 100MP T2I dataset curated with a five-stage automated pipeline which ensures data excellence. Based on PixVerve-95K, we explore distinct training schemes to extend current T2I foundation models to native 100MP generation. Finally, our proposed PixVerve-Bench tailored for UHR scenarios further provides reliable feedback for model evaluation and selection.

\textbf{Limitations and broader impact}. As a work primarily contributing a novel dataset and evaluation benchmark, there remains significant scope for investigating UHR-specific architectural adaptations and more efficient, robust training recipes. And compared to existing general-purpose T2I datasets, the corpus size of PixVerve-95K remains still limited, though our highly scalable curation process. Considering broader impacts, while photorealistic UHR image generation greatly empowers real-world applications, the extreme realism poses heightened ethical risks concerning the proliferation of misinformation and the misuse of AI-generated content. We emphasize vigilance regarding these societal implications within the research community and advocate for multi-dimensional regulations and technical response solutions.

\bibliography{april_aigc}

\clearpage
\newpage
\beginappendix

\renewcommand\thefigure{A\arabic{figure}}
\renewcommand\thetable{A\arabic{table}}  
\renewcommand\theequation{A\arabic{equation}}
\setcounter{equation}{0}
\setcounter{table}{0}
\setcounter{figure}{0}
\appendix
\label{sec: appendix}

The appendix presents the following sections to strengthen the main manuscript:

\begin{itemize}
    \item[—] \textbf{\cref{sec: appendix_data_details}} provides implementation details of flatness detection.
    \item[—] \textbf{\cref{sec: appendix_dataset_analysis}} provides a further frequency-domain analysis to confirm the quality of \dataset.
    \item[—] \textbf{\cref{sec: appendix_license}} provides a detailed clarification on the licensing for our proposed dataset to ensure transparency and ethical compliance.
    \item[—] \textbf{\cref{sec: appendix_dataset_samples}} presents more qualitative samples in \dataset.
    \item[—] \textbf{\cref{sec: appendix_train_infer}} provides more training and inference details of different approaches.
    \item[—] \textbf{\cref{sec: appendix_metric}} provides additional details on metrics of PixVerve-Bench, including specific evaluation procedures, scoring formulas, and human alignment analysis.
    \item[—] \textbf{\cref{sec: appendix_eval}} provides more quantitative and qualitative results of model performance.
    \item[—] \textbf{\cref{sec: appendix_compression}} provides benchmark results of image compression using our 100MP images.
    \item[—] \textbf{\cref{sec: appendix_prompts}} shows the prompts used for MLLM evaluation in PixVerve-Bench.
\end{itemize}

\section{Implementation Details of Flatness Detection} \label{sec: appendix_data_details}
In the stage of preliminary data purification (\cref{sec: purification}), we conduct a flatness detection procedure based on the Sobel variance. Each image in the raw data pool is first converted to grayscale and partitioned into $240 \times 240$ non-overlapping patches. We then compute the gradient magnitude $G_{mag} = \sqrt{G_x^2 + G_y^2}$ for each patch, using Sobel operators with a kernel size of 3. A patch is categorized as ``textureless'' if the variance of its gradient magnitude falls below a predefined threshold of 750, and an image is discarded if the proportion of textureless patches exceeds 97.5$\%$. Notably, both thresholds are intentionally conservative and determined empirically through manual visual audits, which ensures the efficient elimination of overly flat images while still preserving legitimate low-texture content.

\section{Further Analysis of PixVerve-95K Dataset} \label{sec: appendix_dataset_analysis}
\begin{figure}[t]
    \centering
    \includegraphics[width=0.875\linewidth]{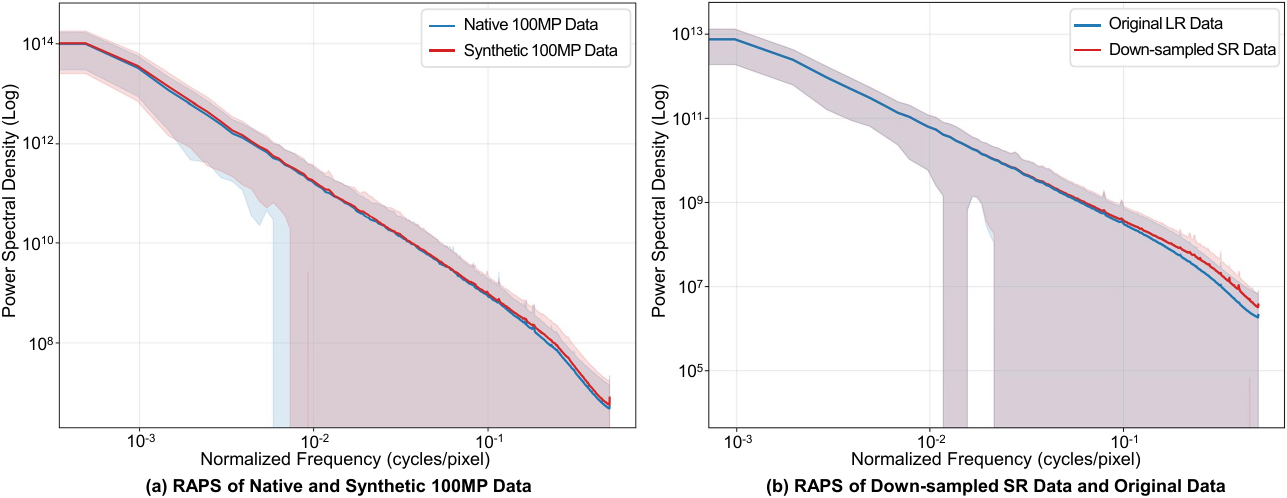}
    \caption{RAPS comparisons.}
    \label{fig: appendix_raps}
\end{figure}

We conduct an extended frequency-domain quality analysis of our PixVerve-95K dataset, utilizing the Radially Averaged Power Spectrum (RAPS). This analysis intuitively presents the power distribution across spatial frequencies, providing insights into the realism of synthesized high-frequency textures. As illustrated in \cref{fig: appendix_raps} (a), the power spectral density of our synthetic 100MP data closely matches the native distribution across the entire frequency range; notably, no significant energy attenuation is observed in the high-frequency regime, confirming the micro-texture fidelity of our rigorously preserved synthetic data. Furthermore, we analyze the consistency between down-sampled SR images and their original low-resolution (LR) counterparts. The alignment of the power spectrum shown in \cref{fig: appendix_raps} (b) indicates that our data curation pipeline maintains global structural consistency and adheres to the underlying distribution of the original images. Overall, this analysis further substantiates that our PixVerve-95K provides high-quality, ultra-high-resolution data, ensuring its reliability for downstream large-scale generative modeling.

\section{Licensing and Dataset Release} \label{sec: appendix_license}
This section provides a detailed account of the licensing for our proposed dataset. As illustrated in the main manuscript, a significant portion of the images was sourced from Pexels~\cite{pexels} and Unsplash~\cite{unsplash}. These platforms operate under permissive licenses that grant broad permissions for downloading, using, and modifying visual content for both commercial and non-commercial purposes without financial obligation. Additionally, a subset is from Aesthetic-Train-V2~\cite{Aesthetic-Train-V2} and UltraHR-100K~\cite{UltraHR-100K}, which are governed by the MIT License and the Creative Commons Attribution-NonCommercial 4.0 (CC BY-NC 4.0) License, respectively. Both frameworks permit the utilization of data for non-commercial research purposes. Therefore, we affirm that our dataset is in full compliance with current copyright laws and privacy regulations, and this dataset is released under the CC BY-NC 4.0 license to prevent unauthorized commercial exploitation.

\section{Qualitative Samples in PixVerve-95K Dataset} \label{sec: appendix_dataset_samples}
\begin{figure}[htbp!]
    \centering
    \includegraphics[width=0.86\linewidth]{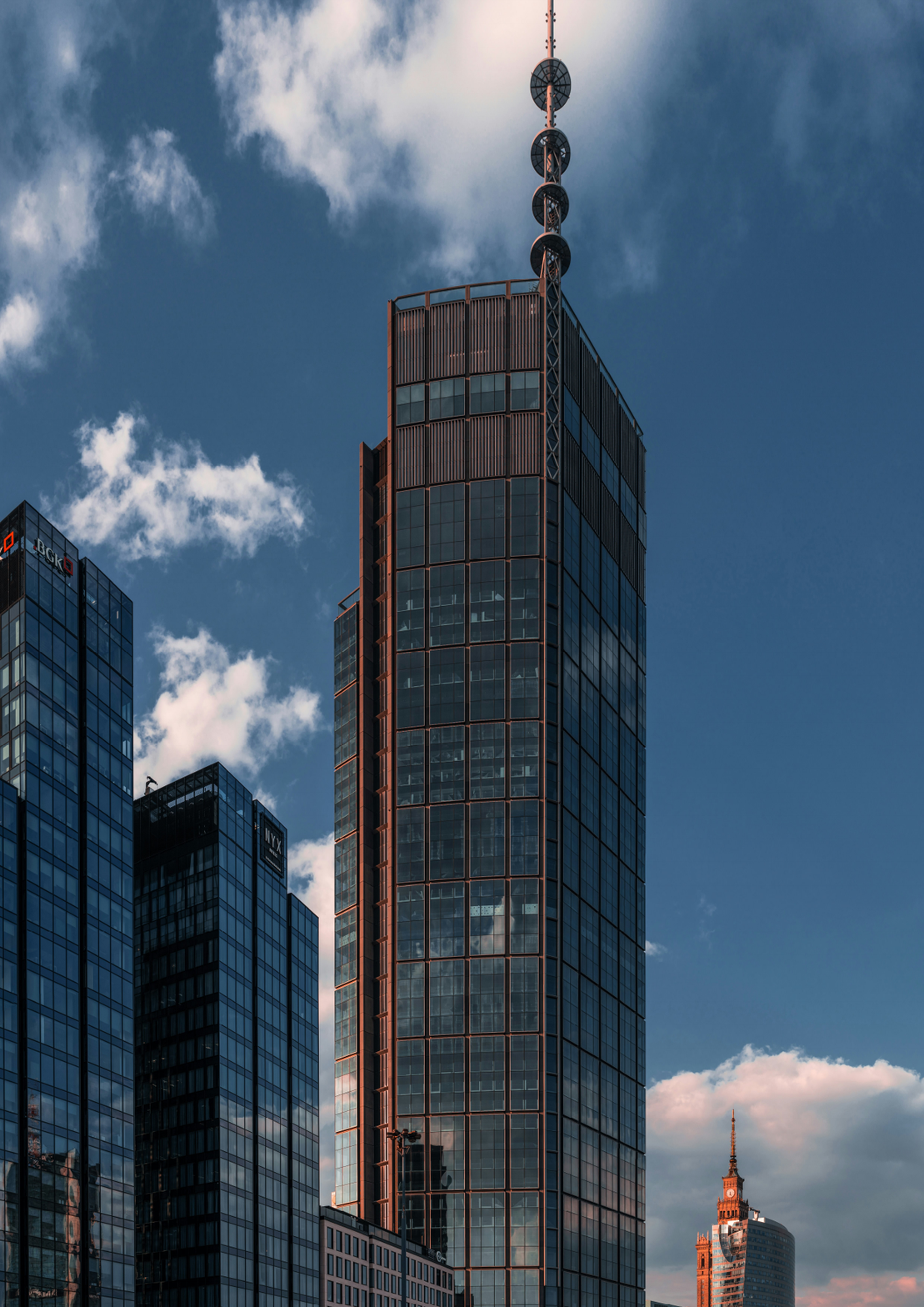}
    \caption{One qualitative 15289$\times$8600 sample in our PixVerve-95K dataset.}
    \label{fig: appendix_sample1}
\end{figure}

\begin{figure}[htbp!]
    \centering
    \includegraphics[width=0.86\linewidth]{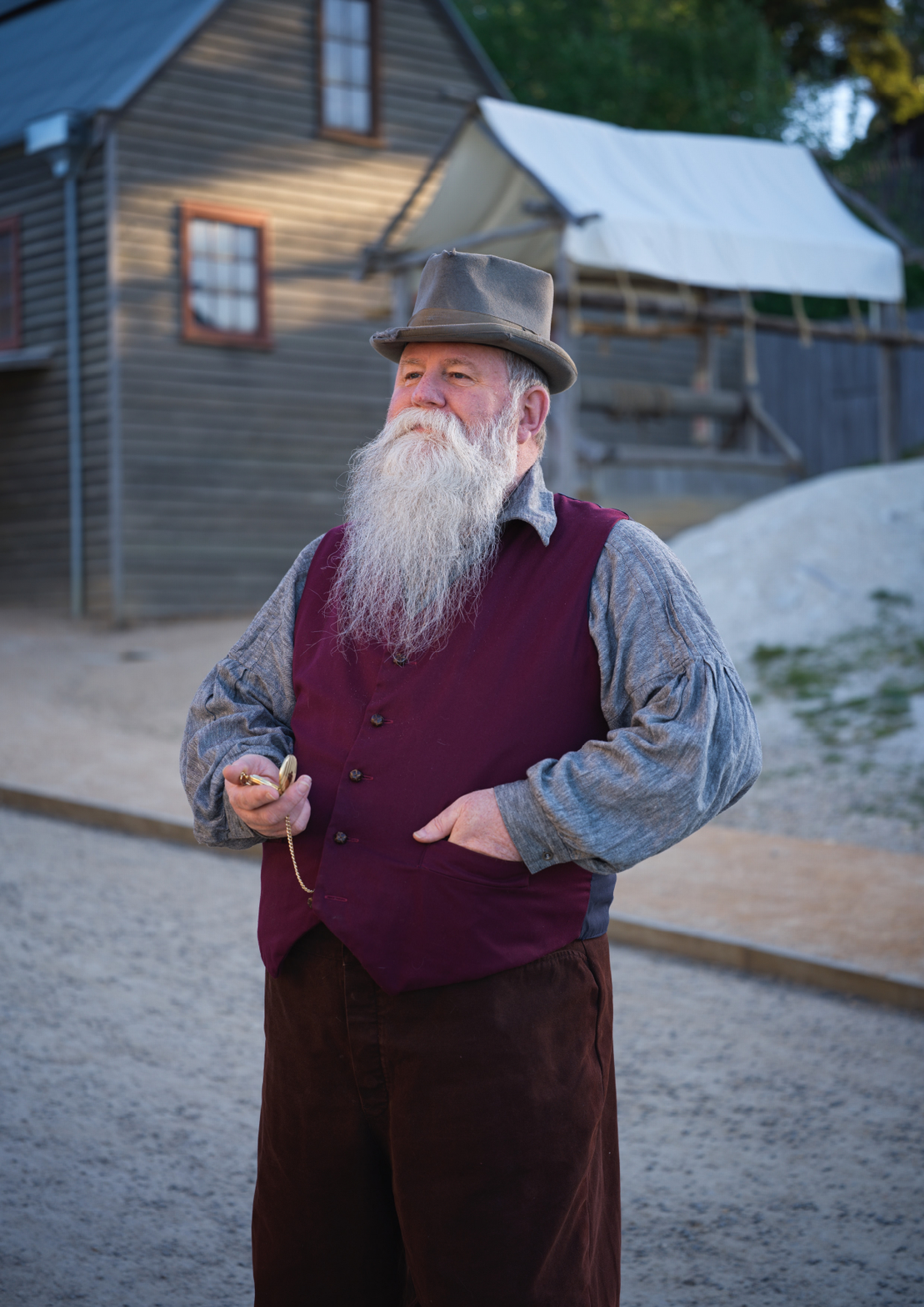}
    \caption{One qualitative 11656$\times$8742 sample in our PixVerve-95K dataset.}
    \label{fig: appendix_sample2}
\end{figure}

\cref{fig: appendix_sample1} and \cref{fig: appendix_sample2} show two qualitative samples in our PixVerve-95K dataset. Best viewed zoomed-in.

\section{More Training and Inference Details} \label{sec: appendix_train_infer}
\subsection{Training Details} \label{sec: appendix_train}
\begin{wraptable}{r}{0.55\textwidth}
    \centering
    \vspace{-0.75em}
    \caption{Training details of different schemes.}
    \label{tab: appendix_training_details}
    \renewcommand{\arraystretch}{1.15}
    \resizebox{1.0\linewidth}{!}{
        \begin{tabular}{ccc}
            \toprule
            \textbf{Training Resolution} & \textbf{Number of Epochs} & \textbf{Total H20 GPU Hours} \\
            \hline
            \rowcolor{blue_tab} \multicolumn{3}{c}{\textbf{\textit{Scheme} $\mathrm{I}$ (Full)}} \\
            4K & 6 & 21,888 \\
            8K & 0.25 & 10,752 \\
            \hline
            \rowcolor{blue_tab} \multicolumn{3}{c}{\textbf{\textit{Scheme} $\mathrm{I}$ (LoRA)}} \\
            4K & 3 & 10,368 \\
            8K & 0.25 & 10,752 \\
            10K & 0.32 & 28,416 \\
            \hline
            \rowcolor{blue_tab} \multicolumn{3}{c}{\textbf{\textit{Scheme} $\mathrm{II}$}} \\
            4K & 3 & 9,216 \\
            \hline            
            \rowcolor{blue_tab} \multicolumn{3}{c}{\textbf{\textit{Scheme} $\mathrm{III}$}} \\
            4K & 4.3 & 8,448 \\
            8K & 2.9 & 18,432 \\
            10K & 1.7 & 23,040 \\
            \toprule
        \end{tabular}
    }
    \vspace{-1em}
\end{wraptable}

For each training scheme, we report the fine-tuning epochs at each resolution scale and the corresponding NVIDIA H20 GPU hours in \cref{tab: appendix_training_details}. GPU hours are computed as the product of the number of NVIDIA H20 GPUs used and the wall-clock training time. Certain schemes do not complete the full three-stage training process since they have exhibited unsatisfactory performance at low resolutions. These computational details are provided to ensure the reproducibility of our main experimental results.

For \textit{Scheme $\mathrm{I}$}, we consider both full-parameter fine-tuning and LoRA for parameter-efficient fine-tuning. Specifically, the full-parameter trained 8K model is initialized from the 4K checkpoint trained for 3 epochs, and is further fine-tuned for 0.25 epochs at 8K resolution. All LoRA models at different resolutions are independently fine-tuned from the same base model, FLUX.2-klein-base-4B, rather than being initialized from lower-resolution LoRA checkpoints. For \textit{Scheme $\mathrm{II}$}, we report the cost of full-parameter fine-tuning with window-attention retrofitting at 4K resolution. For \textit{Scheme $\mathrm{III}$}, the training also follows a progressive curriculum: the 8K model is initialized from the 4K checkpoint, while the 10K model is initialized from the 8K checkpoint after 1 epoch of training.

\subsection{Inference Details}
\begin{wraptable}{r}{0.55\textwidth} 
    \centering 
    \caption{Inference details of different methods.}
    \label{tab: appendix_inference_details}
    \renewcommand{\arraystretch}{1.15}
    \resizebox{1.0\linewidth}{!}{ 
        \begin{tabular}{ccc} 
            \toprule 
            \textbf{Inference Resolution} & \textbf{Number of GPUs} & \textbf{Inference time (s)} \\ 
            \hline 
            \rowcolor{blue_tab} \multicolumn{3}{c}{\textbf{\textit{Scheme} $\mathrm{I}$ (Full)}} \\ 
            4K & 8 & 103 \\ 
            8K & 8 & 1,234 \\
            \hline 
            \rowcolor{blue_tab} \multicolumn{3}{c}{\textbf{\textit{Scheme} $\mathrm{I}$ (LoRA)}} \\ 
            4K & 8 & 103 \\
            8K & 8 & 1,252 \\
            10K & 8 & 2,977 \\
            \hline 
            \rowcolor{blue_tab} \multicolumn{3}{c}{\textbf{\textit{Scheme} $\mathrm{II}$}} \\ 
            4K & 8 & 71 \\
            \hline            
            \rowcolor{blue_tab} \multicolumn{3}{c}{\textbf{\textit{Scheme} $\mathrm{III}$}} \\ 
            4K & 1 & 58 \\ 
            8K & 1 & 70 \\ 
            10K & 1 & 88 \\ 
            \hline            
            \rowcolor{blue_tab} \multicolumn{3}{c}{\textbf{DemoFusion}~\cite{DemoFusion}} \\ 
            4K & 1 & 945 \\
            8K & 1 & 6,366 \\
            10K & 1 & 13,689 \\
            \toprule 
        \end{tabular} 
    } 
    \vspace{-1em} 
\end{wraptable}

In this section, we report the per-sample inference cost of different methods in \cref{tab: appendix_inference_details}. The number of GPUs denotes the number of devices jointly used to generate one sample, rather than the batch size, and the inference time is measured as wall-clock latency in seconds.

For \textit{Scheme $\mathrm{I}$}, as we expect, full fine-tuning and LoRA fine-tuning show nearly identical inference latency, since LoRA changes the adaptation parameterization but does not alter the dominant full-attention computation over the high-resolution latent grid. Therefore, the latency of both variants grows rapidly as the inference resolution increases. Increasing the resolution from 4K to 8K raises the inference time from 103s to over 1,200s, and the LoRA variant requires nearly 3,000s at 10K. Moreover, all these runs require 8 GPUs for generating a single sample, revealing the substantial memory and deployment cost of full-attention latent diffusion at the 100MP scale.

\textit{Scheme $\mathrm{II}$} reduces the 4K inference time from 103s to 71s under the same 8-GPU setting, which verifies that the window-attention retrofitting effectively lowers the attention cost while retaining compatibility with the pretrained latent diffusion backbone. However, it still inherits the multi-GPU inference requirement of the latent-space pipeline, making it an efficiency improvement rather than a complete solution to the deployment bottleneck.

In contrast, \textit{Scheme $\mathrm{III}$} exhibits the most favorable inference behavior. It runs on a single GPU card at all evaluated resolutions, with inference times of 58s, 70s, and 88s at 4K, 8K, and 10K, respectively. The nearly flat latency derives from the adaptive patch-size design, which controls the transformer sequence length as the image resolution increases. Compared to the representative training-free method, DemoFusion, \textit{Scheme $\mathrm{III}$} is $16.3\times$, $90.9\times$, and $155.6\times$ faster at 4K, 8K, and 10K, respectively.

Overall, these results reveal a clear efficiency and deployability trade-off. Full-attention LDM fine-tuning preserves the original computation of the foundation model but becomes prohibitively expensive at ultra-high resolutions. Window-attention retrofitting provides a useful intermediate point by reducing latent attention cost. Patch-based pixel diffusion, although relying on coarser patch-level representations at higher resolutions, is the only explored scheme that enables native 100MP image generation with a single GPU card and sub-minute latency.

\section{Additional Details on Metrics of PixVerve-Bench} \label{sec: appendix_metric}
\subsection{GLCM Score}
\begin{figure}[t]
    \centering
    \includegraphics[width=1\linewidth]{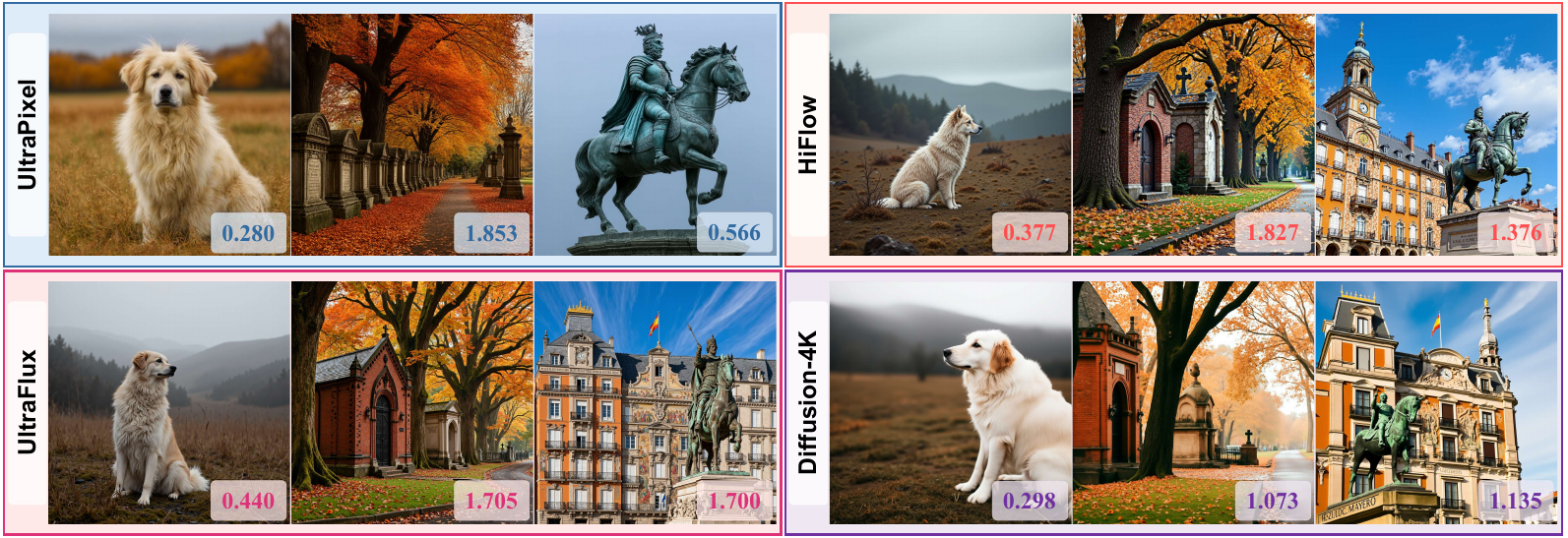}
    \caption{\textbf{Case visualization of the GLCM Score.} We present representative 4096$\times$4096 images generated by UltraPixel, HiFlow, UltraFlux, and Diffusion-4K along with their corresponding scores, demonstrating that higher scores indicate richer texture. Prompts are from PixVerve-Bench.}
    \label{fig: appendix_glcm}
\end{figure}

In this section, we provide the detailed computation procedure of the GLCM Score, which is introduced to quantitatively evaluate the texture granularity in PixVerve-Bench. The GLCM Score is computed by first quantizing the grayscale intensities of the generated image into 64 levels and partitioning it into a set of $64 \times 64$ non-overlapping local patches $\{p_1, p_2, \dots, p_P\}$. For each patch $p_i$, a normalized Gray Level Co-occurrence Matrix~\cite{GLCM} $G_{p_i}$ is constructed across multiple predefined distances $\delta \in \{1, 2, 3, 4\}$ and orientations $\theta \in \{0, \frac{\pi}{4}, \frac{\pi}{2}, \frac{3\pi}{4}\}$ to capture spatial correlations. Subsequently, the detail richness of $p_i$ is measured via the average Shannon entropy~\cite{Shannon}, which is calculated over all spatial parameters and denoted as $H(G_{p_i})$. Finally, the GLCM Score $S$ for the entire image is defined as the arithmetic mean of the entropy values across all $P$ patches:
\[S = \frac{1}{P} \sum_{i=1}^{P} H(G_{p_i}),\]
providing an objective statistical assessment of the micro-structural complexity which is essential for UHR image generation. The case visualization of the GLCM Score is shown in \cref{fig: appendix_glcm}.

\subsection{Multi-scale Fidelity Index (MSFI)}
In this section, we provide the detailed procedure and formulas of computing the Multi-scale Fidelity Index (MSFI).

\noindent \textbf{Taxonomy of evaluation dimensions.}
As described in the main manuscript, the MSFI systematically assesses UHR image fidelity across two complementary dimensions, spanning nine fine-grained sub-dimensions that each target a distinct artifact category. \cref{tab: appendix_msfi} summarizes the evaluation dimensions, sub-dimensions, corresponding descriptions, and criteria.

\begin{figure}[t]
    \centering
    \includegraphics[width=0.735\linewidth]{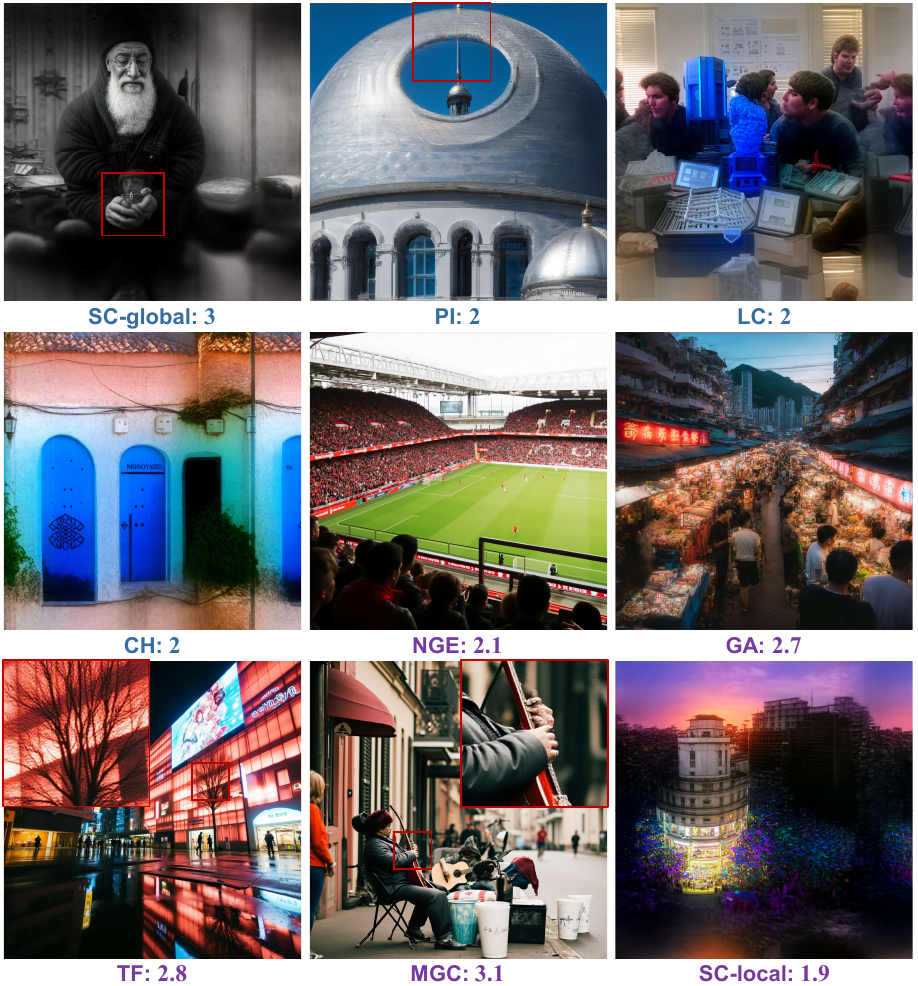}
    \caption{\textbf{Case visualization of the MSFI.} We present typical examples of low-quality generations across the nine sub-dimensions, accompanied by their corresponding scores. \textcolor{b_scores_msfi}{Blue scores} denote sub-dimension ratings for global-scale fidelity, while \textcolor{p_scores_msfi}{purple scores} indicate those for local-scale fidelity, which are the arithmetic means of scores across ten sampled patches. SC-global, PI, LC, CH, NGE, GA, TF, MGC, and SC-local denote the nine MLLM-as-a-judge sub-dimensions described in \cref{tab: appendix_msfi}, respectively. Red bounding boxes are utilized to highlight specific degradations identified within the examples.}
    \label{fig: appendix_msfi}
\end{figure}

\noindent \textbf{Detailed evaluation procedures.}
For each predefined sub-dimension, we instruct Qwen3.5-35B-A3B~\cite{Qwen3.5} to assign a score on a five-point scale reflecting artifact severity. Specifically, for global-scale assessment, the MLLM evaluates the complete generated image resized to a standardized resolution. For local-scale assessment, we adopt the hybrid strategy used in \cref{sec: filtering} to sample ten representative local patches per image, with the patch size set to 512 $\times$ 512 for images below 8K resolution and 1024 $\times$ 1024 otherwise; during each scoring round, the target local patch is evaluated at its native resolution, with the down-sampled complete image provided for contextual reference and the patch's relative spatial coordinates explicitly specified in the prompt; finally, the local-scale fidelity score of each image is derived by averaging the ratings assigned to the ten sampled patches. Furthermore, we provide detailed definitions for each sub-dimension and descriptions for each score level to ensure that the model maintains consistent criteria across multiple rounds of evaluations. The case visualization of each sub-dimension is shown in \cref{fig: appendix_msfi}. The prompts used for global-scale and local-scale fidelity evaluation are provided in \cref{sec: appendix_prompts}.

\begin{table}[t]
    \centering
    \caption{\textbf{Description and evaluation criteria for each sub-dimension of the MSFI.} We detail the hierarchical framework designed to assess the fidelity of UHR images across dual scales.}
    \label{tab: appendix_msfi}
    \small
    \setlength\tabcolsep{5pt}
    \begin{tabularx}{\textwidth}{@{} ll L @{}}
    \toprule
    \textbf{Dimension} & \textbf{Sub-dimension} & \textbf{Description and Evaluation Criteria} \\ 
    \midrule

    \multirow{12}{*}{\textbf{Global-scale}} 
    & Structural Coherence & Evaluates the physical plausibility of overall spatial arrangements and global geometric integrity of entities, ensuring anatomical correctness (\eg, no missing or redundant limbs). \\ \cmidrule(l){2-3}
    & Perspective Integrity & Assesses whether vanishing points and the relative scale of objects at varying depths conform to perspective principles, identifying any geometric distortions against common sense. \\ \cmidrule(l){2-3}
    & Lighting Consistency & Inspects global illumination for naturalism, checking for consistent light direction and the absence of artificial luminance gradients or disjointed shading. \\ \cmidrule(l){2-3}
    & Color Harmony & Examines chromatic transitions for smoothness, checking for quantization artifacts (\eg, color banding) or unnatural boundary blurring between distinct color blocks. \\ 
    \midrule

    \multirow{14}{*}{\textbf{Local-scale}} 
    & Noise \& Grain Existence & Detects stochastic high-frequency chroma noise and unnatural graininess within local patches that deviate from the expected sensor behavior. \\ \cmidrule(l){2-3}
    & Generative Artifacts & Scrutinizes local patches for typical synthesis flaws, including checkerboard patterns, aliasing, and edge halos. \\ \cmidrule(l){2-3}
    & Texture Fidelity & Differentiates between realistic surface roughness and ``plastic-like'' oversmoothing, ensuring natural materials exhibit stochastic randomness rather than mechanical repetition. \\ \cmidrule(l){2-3}
    & Micro-geometry Coherence & Analyzes the continuity of local contours at the pixel level, penalizing unacceptable jitter, ``staircase'' effects, or jagged edges in high-contrast regions. \\ \cmidrule(l){2-3}
    & Sharpness Consistency & Validates the consistency of the focal plane, ensuring that areas within the same depth of field maintain uniform clarity without abnormal, localized blurring. \\ 

    \bottomrule
    \end{tabularx}
\end{table}

\noindent \textbf{Unified scoring formulation.}
For a given evaluation dimension $D$ (global-scale or local-scale), let the set of scores for different sub-dimensions be $\{s_1, s_2, ..., s_{n_D}\}$ with corresponding weights $\{w_1, w_2, ..., w_{n_D}\}$. The score for dimension $D$ is defined as:
\[S_D=\frac{\Sigma_{i=1}^{n_D}w_i \cdot s_i}{\Sigma_{i=1}^{n_D}w_i}.\]
The weights are determined through a user study, where participants provided importance ratings (1-5) for all sub-dimensions. These ratings were averaged, rounded to the nearest integer, and applied directly as the weights. The overall MSFI for image $I$ is given by:
\[\text{MSFI}(I)=S_{global}(I)+w_l \cdot \frac{1}{10}\Sigma_{i=1}^{10}S_{local}(I,i),\]
where $S_{global}(I)$ and $S_{local}(I,i)$ denote the aggregated global fidelity score of $I$ and the local fidelity score of the $i^{th}$ patch sampled from $I$. Notably, we set the weighting factor $w_l$ as $\frac{S_{global(I)}}{5}$, claiming that global structural integrity is a fundamental prerequisite for microscopic realism. This formulation effectively penalizes ``structurally incoherent'' images that might otherwise attain misleadingly high scores due to sharp local textures. Consequently, the MSFI ranges from $1.2$ to $10$, where a score approaching $10$ signifies superior multi-scale fidelity.

\subsection{Instance-centric Compliance Score (ICS)}
In this section, we provide the specific evaluation procedure and scoring formulas of the Instance-centric Compliance Score (ICS), which is proposed to quantify how precisely visual instances adhere to complex textual descriptions. We implement the ICS evaluator leveraging the Qwen3.5-35B-A3B model~\cite{Qwen3.5}.

\noindent \textbf{Evaluation dimensions.}
The assessment of the ICS is performed across three distinct dimensions, each scored by MLLM on a ten-point scale based on specific rubrics:
\begin{itemize}[leftmargin=1em]
    \item \textbf{Instance Existence Verification (IEV):} This dimension serves as the gatekeeper, identifying whether all primary and secondary instances specified in the long caption are present. It focuses purely on presence or absence rather than quality.
    \item \textbf{Appearance Attribute Alignment (AAA):} For identified instances, this dimension evaluates whether the corresponding visual attributes (\eg, color, texture, material, shape, \etc) are compliant with the textual description.
    \item \textbf{Spatial Relation Accuracy (SRA):} This dimension assesses whether the relative positioning (\eg, ``left of'', ``behind'', ``in the foreground'', \etc) and the logical perspective are accurately depicted in the synthesized images.
\end{itemize}
We also provide detailed descriptions for each score level to ensure consistent criteria across multiple rounds of evaluations. The prompt used for instance-centric compliance evaluation is provided in \cref{sec: appendix_prompts}.

\noindent \textbf{Unified scoring formulation.}
For each generated image, we instruct the MLLM to evaluate the aforementioned three dimensions, yielding a set of raw scores denoted as $\{S_{IEV}, S_{AAA}, S_{SRA}\}$. Considering the hierarchical dependency where appearance attributes and spatial relations are contingent upon the existence of the instances themselves, we employ a gated weighted average strategy to synthesize these three-dimensional scores into the final ICS:
\[\text{ICS} = \sqrt{\frac{S_{IEV}}{10}} \times \left( \alpha \cdot S_{AAA} + \beta \cdot S_{SRA} \right),\]
where the term $\sqrt{S_{IEV}/10}$ acts as a penalty for instance omissions. In our implementation, we set $\alpha = 0.6$ and $\beta = 0.4$ to prioritize appearance attribute fidelity. Consequently, an ICS nearing 10 indicates superior instance-centric compliance, while a low score reflects either obvious entity omissions or a misalignment of visual details. 

\subsection{Human Alignment for Metric Validation}
To validate that our proposed automated MSFI and ICS align with human-centric preferences, this section provides a quantitative analysis. We select four representative text-to-image models, $\mathcal{M}=\{M_A, M_B, M_C, M_D\}$, to generate 4K and 8K images. For each resolution, 30 unique prompts are utilized, resulting in a total of $C_4^2\times60=360$ pair-wise comparison sets. We recruit 8 participants to conduct this human preference study, each tasked with evaluating 90 image pairs to ensure that every pair receives two independent annotations. Participants are provided with detailed definitions and illustrative examples for two evaluation dimensions: \textit{Image Fidelity} and \textit{Instance-centric Semantic Alignment}, which correspond to the MSFI and ICS, respectively. For each comparison pair $(I_i, I_j)$ generated by $M_i$ and $M_j$, annotators are instructed to indicate a preference or label the pair as ``indistinguishable''. Let $s_{i,j}$ be the score assigned to model $M_i$ in a single comparison:
$$s_{i,j} = 
\begin{cases} 
1.0 & \text{if } M_i \text{ is preferred,} \\
0.5 & \text{if indistinguishable,} \\
0 & \text{if } M_j \text{ is preferred.}
\end{cases}$$
For each evaluation dimension, the final human preference score for each model is computed as the total score divided by the number of comparisons. We observe that the performance rankings of the four models yielded by our proposed MSFI and ICS perfectly match the rankings derived from human preference scores. This consistency demonstrates that the MSFI and ICS are well-aligned with human subjective judgments in distinguishing the quality and alignment capabilities of different T2I models.

\section{More Quantitative and Qualitative Results} \label{sec: appendix_eval}
\subsection{More Quantitative Results}
\cref{tab: detailed_performance} details the performance of different methods on the MSFI. SC-global, PI, LC, CH, NGE, GA, TF, MGC, and SC-local denote the nine MLLM-as-a-judge sub-dimensions described in \cref{tab: appendix_msfi}, respectively.
\begin{table}[t]
    \centering
    \caption{\textbf{Detailed MSFI comparison across various UHR scales, dimensions, and sub-dimensions.} -- indicates complete failures such as producing meaningless textures or black images, which are not applicable to the semantics-agnostic MSFI. Higher values indicate better performance.}
    \label{tab: detailed_performance}
    \setlength{\tabcolsep}{3.25pt} 
    \renewcommand{\arraystretch}{1.15}
    \resizebox{\textwidth}{!}{
        \begin{tabular}{cc|cc|ccccccccc}
            \toprule
            \multirow{3}{*}{\makecell{\textbf{Resolution} \\ \textbf{(height $\times$ width)}}} & \multirow{3}{*}{\textbf{Method}} & \multicolumn{2}{c|}{\textbf{Dimension Performance}} & \multicolumn{9}{c}{\textbf{Sub-dimension Performance}} \\
            \cmidrule(lr){3-4} \cmidrule(lr){5-13}
            & & \makecell{\textbf{Global} \\ \textbf{Fidelity}} & \makecell{\textbf{Local} \\ \textbf{Fidelity}} & SC-global & PI & LC & CH & NGE & GA & TF & MGC & SC-local \\
            \midrule
            \multirow{13}{*}{\makecell{\textbf{4K} \\ \textbf{(4096 $\times$ 4096)}}} 
            & FLUX.2-klein-base-4B & 3.955 & 4.366 & 3.485 & 4.260 & 4.240 & 4.385 & 4.326 & 4.381 & 4.332 & 4.409 & 4.426 \\
            & Qwen-Image           & 4.343 & 4.547 & 4.010 & 4.470 & 4.620 & 4.710 & 4.546 & 4.560 & 4.515 & 4.573 & 4.583 \\
            & L2P           & 4.572 & 2.539 & 4.255 & 4.820 & 4.785 & 4.780 & 2.508 & 2.550 & 2.531 & 2.556 & 2.556 \\
            & DemoFusion           & 4.237 & 4.731 & 3.685 & 4.510 & 4.665 & 4.780 & 4.786 & 4.713 & 4.704 & 4.724 & 4.776 \\
            & LinFusion            & 4.410 & 4.264 & 3.935 & 4.685 & 4.750 & 4.845 & 4.315 & 4.291 & 4.202 & 4.295 & 4.301 \\
            & HiFlow               & 4.471 & 4.848 & 4.180 & 4.910 & 4.915 & 4.930 & 4.889 & 4.831 & 4.828 & 4.847 & 4.885 \\
            & UltraPixel           & 4.688 & 4.761 & 4.315 & 4.945 & 4.950 & 4.975 & 4.783 & 4.762 & 4.741 & 4.766 & 4.786 \\
            & UltraFlux            & 4.669 & 4.330 & 4.260 & 4.945 & 4.965 & 4.980 & 4.314 & 4.341 & 4.312 & 4.354 & 4.351 \\
            & Diffusion-4K         & 4.577 & 4.151 & 4.165 & 4.860 & 4.875 & 4.885 & 4.125 & 4.171 & 4.121 & 4.186 & 4.187 \\
            \rowcolor{blue_tab} \cellcolor{white} & FLUX.2-$\mathrm{I}$ (Full)                 & 4.561 & 4.762 & 4.120 & 4.860 & 4.870 & 4.905 & 4.760 & 4.765 & 4.747 & 4.770 & 4.786 \\
            \rowcolor{blue_tab} \cellcolor{white} & FLUX.2-$\mathrm{I}$ (LoRA)                 & 4.575 & 4.801 & 4.155 & 4.865 & 4.855 & 4.910 & 4.815 & 4.796 & 4.785 & 4.807 & 4.830 \\
            \rowcolor{blue_tab} \cellcolor{white} & FLUX.2-$\mathrm{II}$ & 4.150 & 4.784 & 3.650 & 4.530 & 4.365 & 4.615 & 4.791 & 4.792 & 4.761 & 4.797 & 4.814 \\
            \rowcolor{blue_tab} \cellcolor{white} & L2P-$\mathrm{III}$                 & 4.493 & 2.764 & 4.165 & 4.815 & 4.660 & 4.665 & 2.750 & 2.768 & 2.753 & 2.782 & 2.785 \\
            \midrule
            
            \multirow{6.5}{*}{\makecell{\textbf{8K} \\ \textbf{(8192 $\times$ 8192)}}}
            & FLUX.2-klein-base-4B & -- & -- & -- & -- & -- & -- & -- & -- & -- & -- & -- \\
            & L2P & 4.558 & 3.317 & 4.240 & 4.830 & 4.735 & 4.770 & 3.216 & 3.361 & 3.292 & 3.377 & 3.358 \\
            & DemoFusion           & 3.647 & 4.550 & 3.075 & 3.770 & 4.110 & 4.430 & 4.763 & 4.615 & 4.313 & 4.647 & 4.735 \\
            & LinFusion            & 4.250 & 4.173 & 3.755 & 4.510 & 4.615 & 4.730 & 4.230 & 4.193 & 4.121 & 4.193 & 4.200 \\
            \rowcolor{blue_tab} \cellcolor{white} & FLUX.2-$\mathrm{I}$ (Full)                 & -- & -- & -- & -- & -- & -- & -- & -- & -- & -- & -- \\
            \rowcolor{blue_tab} \cellcolor{white} & FLUX.2-$\mathrm{I}$ (LoRA)                 & -- & -- & -- & -- & -- & -- & -- & -- & -- & -- & -- \\
            \rowcolor{blue_tab} \cellcolor{white} & L2P-$\mathrm{III}$                 & 3.490 & 2.580 & 3.255 & 4.020 & 3.410 & 3.365 & 2.542 & 2.594 & 2.563 & 2.601 & 2.618 \\
            \midrule
            
            \multirow{4}{*}{\makecell{\textbf{10K} \\ \textbf{(10240 $\times$ 10240)}}}
            & L2P                 & 4.301 & 3.573 & 3.970 & 4.585 & 4.495 & 4.510 & 3.468 & 3.617 & 3.548 & 3.627 & 3.627 \\
            & DemoFusion           & 3.448 & 4.730 & 2.890 & 3.525 & 3.945 & 4.230 & 4.810 & 4.704 & 4.697 & 4.722 & 4.785 \\
            & LinFusion            & 4.139 & 4.090 & 3.620 & 4.400 & 4.545 & 4.640 & 4.145 & 4.113 & 4.041 & 4.107 & 4.108 \\
            \rowcolor{blue_tab} \cellcolor{white} & L2P-$\mathrm{III}$                 & 3.577 & 2.746 & 3.204 & 4.017 & 3.744 & 3.680 & 2.711 & 2.745 & 2.721 & 2.757 & 2.835 \\
            \bottomrule
        \end{tabular}
    }
\end{table}

\subsection{More Qualitative Results}
\begin{figure}[htbp!]
    \centering
    \vspace{1.35em}
    \includegraphics[width=0.835\linewidth]{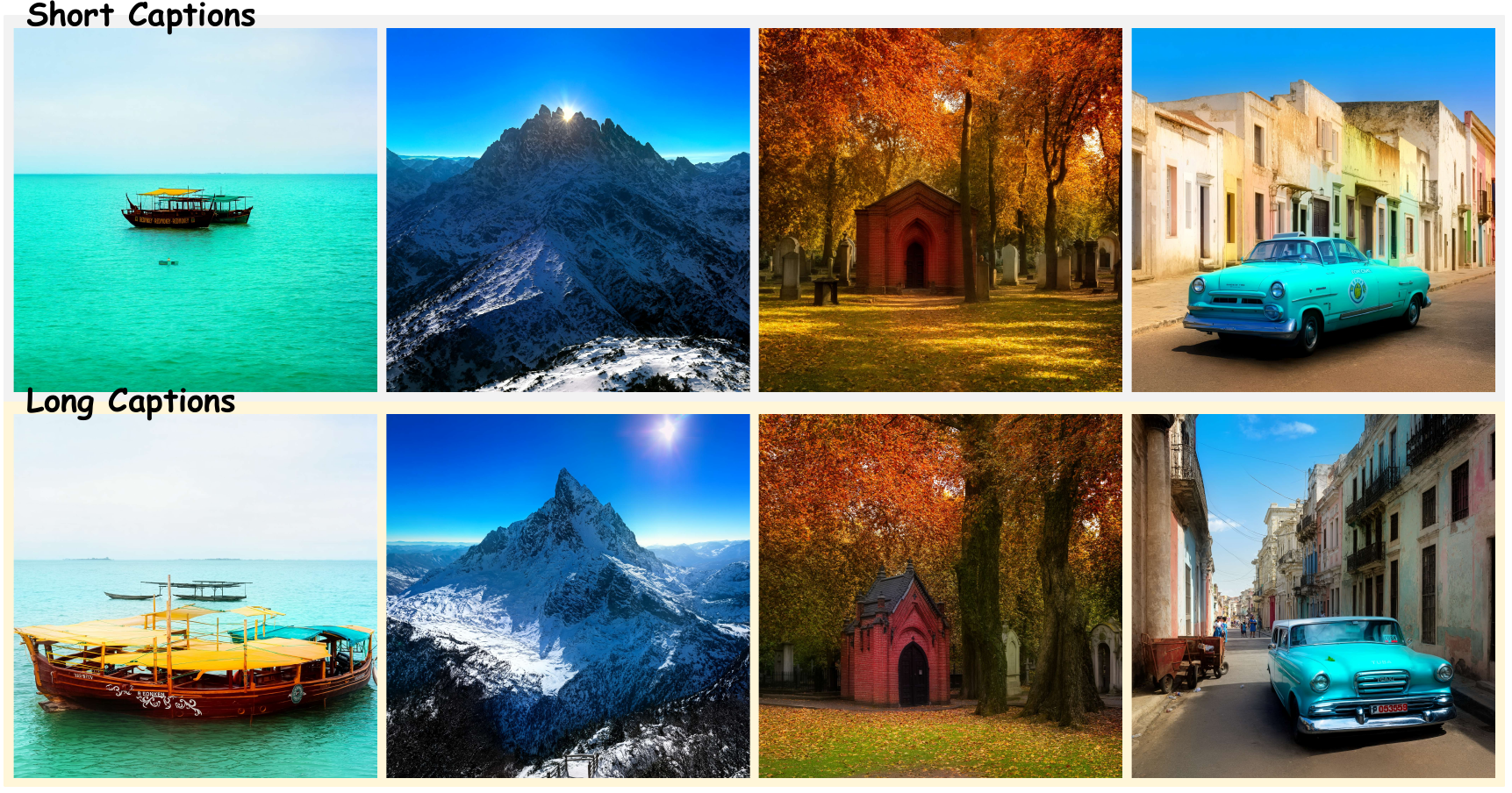}
    \caption{\textbf{Qualitative comparison of 4K image generations under short and long prompting settings.} \textit{Top:} images generated by our FLUX.2-$\mathrm{I}$ (Full) using short captions; \textit{bottom:} generated images using corresponding long captions.}
    \label{fig: appendix_ablation_captions}
\end{figure}

\cref{fig: appendix_ablation_captions} visualizes the qualitative comparison of 4K image generations under short and long prompting settings. All case pairs are generated by our FLUX.2-$\mathrm{I}$ (Full).

\section{Benchmarking for Image Compression at 100MP Scale} \label{sec: appendix_compression}
Following the benchmark specifications in ~\cite{Image-Compression-Benchmark}, we convert 25 images from our dataset into PNM format (a simple image format which stores raw pixels) for standard testing. A total of 20 image compressed formats are evaluated under strict lossless settings. The testing environment consists of a $\text{12}^\text{{th}}$ Gen Intel i7-12700H CPU @2.30 GHz, 16 GB of RAM, and the Windows 11 operating system, and all codecs run in a single thread. The detailed benchmark results, including compressed size, compression time, and decompression time, are summarized in \cref{tab: appendix_compression}.

\begin{table}[htbp!]
    \centering
    \caption{Performance comparison of different lossless image compressed formats at the 100MP scale.}
    \label{tab: appendix_compression}
    \renewcommand{\arraystretch}{1.15}
    \resizebox{1\textwidth}{!}{
    \begin{tabular}{cccc}
        \toprule[0.1em]
        \textbf{Compressed Format} & \textbf{Compressed Size (bytes) $\downarrow$} & \textbf{Compression Time (s) $\downarrow$} & \textbf{Decompression Time (s) $\downarrow$} \\
        \hline
        fNBLI & 2152173140 & 68.343 & 44.276 \\
        NBLI (-g) & 2122763920 & 483.106 & 429.641 \\
        HALIC & 2327457195 & 23.266 & 33.072 \\
        HALICfast & 2501106140 & 16.154 & 25.109 \\
        KVICK & 2467300114 & 60.579 & 43.413 \\
        QIC & 2675297097 & 32.500 & 30.960 \\
        QLIC2 & 2415123224 & 40.452 & 38.795 \\
        LEA & 1898470283 & 772.283 & 870.666 \\
        BMF & 2260500400 & 554.801 & 254.198 \\
        BIM & 2399426455 & 301.153 & 339.631 \\
        QOI & 3685837372 & 28.764 & 22.864 \\
        QOIR & 3421495512 & 28.557 & 26.181 \\
        ZPNG & 3003246621 & 36.658 & 29.638 \\
        PNG (optimizing=False) & 3207101968 & 760.569 & 65.350 \\
        PNG (optimizing=True) & 3138083188 & 5417.594 & 74.167 \\
        JPEG-XL (-q 100 -e 1) & 2765116877 & 77.526 & 107.456 \\
        JPEG-XL (-q 100 -e 2) & 2557728595 & 233.997 & 112.632 \\
        JPEG-XL (-q 100 -e 3) & 2315011289 & 511.114 & 314.402 \\
        WEBP (lossless m1) & 2387603822 & 2536.114 & 75.605 \\
        JPEG-LS & 2969890552 & 234.688 & 199.415 \\
        \toprule[0.1em]
    \end{tabular}
    }
\end{table}

\section{Prompts for MLLM Evaluation} \label{sec: appendix_prompts}
\begin{promptbox}{\small Global-scale Fidelity Evaluation Prompt}
\footnotesize
$\#$ ROLE AND TASK FORMULATION:
\vspace{0.175em}

You are a professional image quality assessment (IQA) and scoring expert specializing in T2I evaluation. Your task is to carefully evaluate the **global-scale fidelity** of the provided synthetic image across 4 dimensions. Please focus on the overall composition and structure, geometric and physical logic, and macro-scale consistency.

\vspace{1em}
$\#$ CRITICAL SCORING RULES (Must Strictly Follow):
\vspace{0.175em}

1. **Objectivity \& Fairness:** Maintain an objective stance throughout the evaluation process and base your judgement on visual evidence with the same standard instead of subjective preference. \\
2. **Focus Solely on Fidelity:** Consider the image category and expected characteristics while avoiding any bias towards the content of the image. Score based on the visual quality and fidelity aspects solely. \\
3. **Independence:** Evaluate each dimension independently without any halo effects. \\
4. **Rigor:** Apply strict criteria and any noticeable artifact should be reflected in the scoring. Maintain a high standard for what constitutes a ``5'' (Excellent).

\vspace{1em}
$\#$ EVALUATION RUBRICS:

\vspace{0.5em}
$\#\#$ 1. Structural Coherence (SC-global) \\
Check whether the geometric structure of the entities is correct, whether there are any missing or redundant limbs, and whether the overall spatial relations are consistent with physical common sense. \\
- 5 (Excellent): Flawless physical logic. Objects are perfectly formed; no missing/extra parts. \\
- 4 (Good): Minor structural oddities that don't distract from the main subject. \\
- 3 (Fair): Noticeable structural errors (e.g., slightly deformed limbs or merged background objects). \\
- 2 (Poor): Obvious physical failures; objects are partially collapsed or mutated. \\
- 1 (Very Poor): Severe structural collapse; chaotic composition; unrecognizable forms.

\vspace{0.5em}
$\#\#$ 2. Perspective Integrity (PI) \\
Check whether the perspective relationship between objects at different distances conforms to the principles of perspective, and whether there is any distorted perspective. \\
- 5 (Excellent): Flawless perspective and geometric projection. Vanishing points and horizon lines align accurately. \\
- 4 (Good): Slight perspective tilt, but geometric projection still feels natural. \\
- 3 (Fair): Distorted depth; objects at different distances feel ``stacked'' or misaligned. \\
- 2 (Poor): Severe geometric warping; architectural lines curve unnaturally or conflict. \\
- 1 (Very Poor): Multiple conflicting vanishing points or warped architectural lines; total perspective failure.

\vspace{0.5em}
$\#\#$ 3. Lighting Consistency (LC) \\
Check whether the overall lighting has consistency with that of a natural image, and whether there are obviously artificial brightness gradients. \\
- 5 (Excellent): Unified light source. Shadows, highlights, and reflections follow ray-tracing logic. \\
- 4 (Good): Consistent lighting, but subtle mismatch in shadow softness or intensity. \\
- 3 (Fair): Ambiguous light source. Some objects appear ``self-lit'' without casting shadows. \\
- 2 (Poor): Contradictory lighting directions. Shadows cast in different ways for nearby objects. \\
- 1 (Very Poor): Complete lighting failure; flat ``sticker-like'' objects with zero interaction with the environment.

\vspace{0.5em}
$\#\#$ 4. Color Harmony (CH) \\
Check whether the overall color transitions are smooth and natural, and whether there are issues such as blurred edges of color blocks and color banding. \\
- 5 (Excellent): Natural color gamut. Smooth gradients; no banding or abnormal color noise. \\
- 4 (Good): High-quality color, though slight over-saturation or other issues in small areas. \\
- 3 (Fair): Visible color banding in gradients. Slight chromatic aberrations. \\
- 2 (Poor): Patchy color blocks; unnatural ``neon'' artifacts or gray/dull patches in vibrant areas. \\
- 1 (Very Poor): Severe color corruption; massive chromatic noise or broken color channels.

\vspace{1em}
$\#$ OUTPUT RULES (Must Strictly Follow):
\vspace{0.175em}

1. You MUST follow a strict 5-point scale and provide a score as an **INTEGER from 1 to 5 only** for each dimension. \\
2. Provide the final output strictly **in a JSON object inside the <json> tag**. \\
3. The <json> block MUST contain ONLY the valid JSON object. No markdown code blocks or extra text. \\
4. Keys: ``SC-global'', ``PI'', ``LC'', ``CH'' represent the scores for the 4 dimensions respectively, and ``reasoning'' is a concise explanation justifying the scores. Ensure the JSON property names are enclosed in double quotes and there are no trailing commas in the JSON object.

\vspace{1em}
$\#$ OUTPUT FORMAT:
\vspace{0.175em}

\texttt{<json> \\
\{\{
\begin{itemize}[leftmargin=1em, label={}, nosep]
    \item ``SC-global'': int,
    \item ``PI'': int,
    \item ``LC'': int,
    \item ``CH'': int,
    \item ``reasoning'': ``A concise explanation justifying the four-dimensional scores.''
\end{itemize}
\}\} \\
</json>}

\vspace{1em}
Now, please evaluate the **global-scale fidelity** of the provided image based on the above criteria and output the scores and reasoning in the specified **JSON** format.
\end{promptbox}

\begin{promptbox}{\small Local-scale Fidelity Evaluation Prompt}
\footnotesize
$\#$ ROLE AND TASK FORMULATION:
\vspace{0.175em}

You are a professional image quality assessment (IQA) and scoring expert specializing in local fine-grained details evaluation. Your task is to carefully evaluate the **local-scale fidelity** of the provided **local patch** of an ultra-high-resolution synthetic image across 5 dimensions. To ensure high-quality evaluation, please adhere to the following guidelines:

\vspace{1em}
$\#$ IMPORTANT CONTEXT:
\vspace{0.175em}

- **Image 1 (First):** This is the **LOCAL PATCH (target for scoring)**. Ignore incomplete objects or composition issues. Focus only on the visual quality of the visible area. \\
- **Image 2 (Second):** This is the **FULL GLOBAL IMAGE (for contextual reference)**. **DO NOT** use this image for direct visual analysis or evaluation. Use this image ONLY for **understanding the original image's theme and global context**. \\
- **Patch Location:** The local patch (Image 1) corresponds to the area defined by the relative coordinates \texttt{\{relative\_coords\}} in the global image (Image 2). The relative coordinates are normalized $[\text{x}_\text{min}, \text{y}_\text{min}, \text{x}_\text{max}, \text{y}_\text{max}]$, where [0,0] is the top-left corner and [1,1] is the bottom-right corner of the global image.

\vspace{1em}
$\#$ CRITICAL SCORING RULES (Must Strictly Follow):
\vspace{0.175em}

1. **Objectivity \& Fairness:** Maintain an objective stance throughout the evaluation process and base your judgement on visual evidence with the same standard instead of subjective preference. \\
2. **Focus Solely on Fidelity:** Consider the image category and expected characteristics while avoiding any bias towards the content of the image. Score based on the visual quality and fidelity aspects solely. \\
3. **Local-to-Global Evaluation:** Evaluate the details in Image 1, and use Image 2 to distinguish between ``intended bokeh/blur'' and ``accidental artifacts''. \\
4. **Coordinates Reference:** Use the rectangular bounding box only to understand the local patch's location in the overall image context, but DO NOT directly compare the local patch to the global image for pixel-level details. \\
5. **Independence:** Evaluate each dimension independently without any halo effects. \\
6. **Rigor:** Apply strict criteria and any noticeable artifact should be reflected in the scoring. Maintain a high standard for what constitutes a ``5'' (Excellent).

\vspace{1em}
$\#$ EVALUATION RUBRICS:
\vspace{0.175em}

Please evaluate the microscopic details and fidelity of the **Local Patch (Image 1)** across the 5 dimensions below, while using the Global Image (Image 2) and the relative coordinates \texttt{\{relative\_coords\}} as reference.

\vspace{0.5em}
$\#\#$ 1. Noise and Grain Existence (NGE) \\
Check whether there is random high-frequency color noise and obvious color graininess in the local patch. \\
- 5 (Excellent): Crystal clean and realistic cinematic grain. Zero digital noise or compression blocks. Grain looks like natural film if present. \\
- 4 (Good): Slight luminance noise, barely visible at 100$\%$ zoom. \\
- 3 (Fair): Noticeable noise or grain that distracts from the details, especially in shadow areas. \\
- 2 (Poor): Heavy salt-and-pepper noise or distracting grain. \\
- 1 (Very Poor): Image details are buried under severe noise or compression corruption.

\vspace{0.5em}
$\#\#$ 2. Generative Artifacts (GA) \\
- 5 (Excellent): No AI-specific artifacts. Details look photo-realistic and like they were captured by a high-end CMOS sensor. \\
- 4 (Good): Minor generative patterns that require close inspection to find. \\
- 3 (Fair): Noticeable AI ``melting'' or ``waxy'' textures where details should be sharp. \\
- 2 (Poor): Hallucinated textures or ``ghosting'' artifacts typical of diffusion models. \\
- 1 (Very Poor): Massive generative collapse; ``AI-soup'' textures.

\vspace{0.5em}
$\#\#$ 3. Texture Fidelity (TF) \\
Check whether the local patch presents plastic-like oversmoothing, and for natural objects such as wood and fabric, whether the texture has randomness rather than mechanical repetition. \\
- 5 (Excellent): Tactile realism. Skin pores, fabric weaves, or surface grit are ultra-sharp and authentic. \\
- 4 (Good): High detail, but slightly ``soft'' or over-regularized texture. \\
- 3 (Fair): Texture is visible but ``flat''; lacks the micro-depth of real-world surfaces. \\
- 2 (Poor): Over-smoothed ``plastic'' look; details are smeared out. \\
- 1 (Very Poor): Completely blurred or ``mushy'' surfaces with zero recognizable texture.

\vspace{0.5em}
$\#\#$ 4. Micro-geometry Coherence (MGC) \\
Check at a local scale whether the lines show unacceptable jitter or jagged edges. \\
- 5 (Excellent): Perfect edge continuity. Fine lines (e.g., hair, wires) are smooth at the pixel level. \\
- 4 (Good): Sharp edges, though very minor aliasing (stair-stepping) visible on diagonals. \\
- 3 (Fair): Jagged edges or slight shimmering; fine lines appear broken in some places. \\
- 2 (Poor): Severe aliasing; pixelated edges; micro-structures look ``broken''. \\
- 1 (Very Poor): Total geometric chaos at the micro-level; edges are unrecognizable.

\vspace{0.5em}
$\#\#$ 5. Sharpness Consistency (SC-local) \\
Check whether there are unnatural blurry patches (i.e., within the same focal plane some areas is clear enough while others are abnormally blurry). \\
- 5 (Excellent): Natural optical sharpness variation consistent with depth of field. No inconsistent blur patches within the same focal plane. \\
- 4 (Good): Slightly soft, but no inconsistent blur patches within the same focal plane. \\
- 3 (Fair): Noticeable inconsistency in sharpness; some areas look artificially sharpened while others are blurry without a natural depth-of-field reason. \\
- 2 (Poor): Obvious sharpness inconsistency; ``cut-and-paste'' feel where the patch looks like it was taken from a different image with different focus. \\
- 1 (Very Poor): Severe sharpness failure; the patch looks like a low-quality thumbnail pasted into the global image.

\vspace{1em}
$\#$ OUTPUT RULES (Must Strictly Follow):
\vspace{0.175em}

1. You MUST follow a strict 5-point scale and provide a score as an **INTEGER from 1 to 5 only** for each dimension. \\
2. Provide the final output strictly **in a JSON object inside the <json> tag**. \\
3. The <json> block MUST contain ONLY the valid JSON object. No markdown code blocks or extra text. \\
4. Keys: ``NGE'', ``GA'', ``TF'', ``MGC'', ``SC-local'' represent the scores for the 5 dimensions respectively, and ``reasoning'' is a concise explanation justifying the scores. Ensure the JSON property names are enclosed in double quotes and there are no trailing commas in the JSON object.

\vspace{1em}
$\#$ OUTPUT FORMAT:
\vspace{0.175em}

\texttt{<json> \\
\{\{
\begin{itemize}[leftmargin=1em, label={}, nosep]
    \item ``NGE'': int,
    \item ``GA'': int,
    \item ``TF'': int,
    \item ``MGC'': int,
    \item ``SC-local'': int,
    \item ``reasoning'': ``A concise explanation justifying the five-dimensional scores.''
\end{itemize}
\}\} \\
</json>}

\vspace{1em}
Now, please evaluate the **local-scale fidelity** of the provided local crop based on the above criteria and output the scores and reasoning in the specified **JSON** format.
\end{promptbox}

\begin{promptbox}{\small Instance-centric Compliance Evaluation Prompt}
\footnotesize
$\#$ ROLE AND TASK FORMULATION:
\vspace{0.175em}

You are a professional image quality assessment and scoring expert specializing in Text-to-Image (T2I) semantic alignment evaluation. Your task is to carefully perform a fine-grained, instance-centric assessment of the provided synthesized image based on its initial detailed long caption. \\
**Input Long Caption:** ``\texttt{\{long\_caption\}}''

\vspace{1em}
$\#$ METRIC DIMENSIONS AND BOUNDARIES:
\vspace{0.175em}

Please evaluate the image across the following three distinct, hierarchical dimensions, strictly adhering to the defined criteria and boundaries: \\
1. **IEV (Instance Existence Verification):** Inspect whether all instances explicitly mentioned in the long caption are present. Focus strictly and solely on presence or absence rather than quality. \\
2. **AAA (Appearance Attribute Alignment):** For each instance that exists, assess whether its visual attributes (color, texture, material, size, shape) align with the description in the long caption. This requires detailed cross-referencing between the caption and the visual content. \\
3. **SRA (Spatial Relation Accuracy):** Evaluate whether the relative positioning (e.g., left/right, top/bottom, foreground/background) and the logical perspective between multiple instances are accurately depicted in the image.

\vspace{1em}
$\#$ CRITICAL SCORING RULES (Must Strictly Follow):
\vspace{0.175em}

1. **Hierarchical Dependence:** **IEV** is the gatekeeper. If any critical instance is missing (IEV below 4), the corresponding AAA and SRA for the image must be penalized accordingly, as attributes and relations cannot exist without the entity. \\
2. **Detail Awareness:** Since this is a high-resolution image evaluation task, you must meticulously scan **the entire canvas**, including corners and background, to identify all mentioned instances and their micro-details. \\
3. **Strict Adherence to Explicit Constraints:** Judge the image ONLY based on what is explicitly stated in the long caption. Do not impose imaginary constraints or personal aesthetic preferences. For any visual aspects NOT mentioned (e.g., specific lighting, background nuances, or artistic style), the generation model is allowed creative autonomy. Do not penalize the model for ``making choices'' where the prompt is silent. \\
4. **Hallucination Penalty:** If the synthesized image contains prominent instances that are NOT mentioned in the long caption and significantly distract from the caption's content (severe hallucination), deduct 1-2 points from **IEV**. \\
5. **No Middle Ground Bias:** Avoid giving 7/10 by default. Be decisive based on the visual evidence. \\
6. **Objectivity \& Fairness:** Maintain an objective stance throughout the evaluation process and base your judgement on visual evidence with the same standard instead of subjective preference.

\vspace{1em}
$\#$ SCORING RUBRICS (10-Point Scale):

\vspace{0.5em}
$\#\#$ 1. IEV (Instance Existence Verification) \\
- **[9-10]:** All instances (primary and secondary) from the long caption are present and clearly identifiable in the image. No significant omissions. \\
- **[7-8]:** Primary instances are present; only minor, non-essential background elements are missing or extremely unclear. \\
- **[5-6]:** At least one primary instance is missing, or multiple instances are severely obscured; but the image still captures the general theme and content of the caption. \\
- **[3-4]:** Multiple primary instances are missing or unrecognizable, significantly detracting from the caption's description. \\
- **[1-2]:** Total mismatch; the image fails to depict the core content of the caption.

\vspace{0.5em}
$\#\#$ 2. AAA (Appearance Attribute Alignment) \\
- **[9-10]:** Perfect alignment. All visible attributes (color, texture, material, size, shape) of the instances matches the caption exactly. \\
- **[7-8]:** Most attributes are correct; slight deviations in secondary details (e.g., shade of color or minor texture mismatch) that do not significantly affect the overall perception. \\
- **[5-6]:** Noticeable attribute misalignment in primary instances (e.g., wrong color or material), but the image still somewhat reflects the caption's content. \\
- **[3-4]:** Severe attribute misalignment; primary instances look totally different from the text description. \\
- **[1-2]:** Complete attribute failure; objects lack any descriptive fidelity or are rendered as generic blobs.

\vspace{0.5em}
$\#\#$ 3. SRA (Spatial Relation Accuracy) \\
- **[9-10]:** All relative positions between instances and depth cues perfectly match the spatial prepositions in the caption. \\
- **[7-8]:** Relative positions are correct, but there are minor errors in scale/perspective or overlapping logic that do not cause major confusion. \\
- **[5-6]:** Noticeable spatial relation errors (e.g., left/right flipped, foreground/background confusion). \\
- **[3-4]:** Obvious spatial relation failures; instances are positioned in a way that contradicts the caption's logic. \\
- **[1-2]:** Chaotic layout; objects are floating or positioned randomly without regard for the caption's logic.

\vspace{1em}
$\#$ OUTPUT RULES (Must Strictly Follow):
\vspace{0.175em}

1. You MUST follow a strict 10-point scale and provide a score as an **INTEGER from 1 to 10 only** for each dimension. \\
2. Provide the final output strictly **in a JSON object inside the <json> tag**. \\
3. The <json> block MUST contain ONLY the valid JSON object. No markdown code blocks or extra text. \\
4. Keys: ``reasoning'', ``IEV'', ``AAA'', and ``SRA''. ``reasoning'' is a concise explanation justifying the scores; ``IEV'', ``AAA'', and ``SRA'' represent the scores for the 3 dimensions respectively. Ensure the JSON property names are enclosed in double quotes and there are no trailing commas in the JSON object.

\vspace{1em}
**Output example for reference (do not copy this exact content, just an example of the structure):** \\
\texttt{<json> \\
\{\{
\begin{itemize}[leftmargin=1em, label={}, nosep]
    \item ``reasoning'': ``The cat and the bench is present but the straw hat is missing; the absence of the hat results in a deduction in IEV. The cat's color and texture are reasonably well-aligned with the caption, though the fur appears more orange than described, leading to a moderate AAA score. The spatial relations are mostly accurate, with the cat sitting on bench and positioned on the left as specified, but the background elements are somewhat jumbled, resulting in a high but not perfect SRA score.'',
    \item ``IEV'': 6,
    \item ``AAA'': 8,
    \item ``SRA'': 9
\end{itemize}
\}\} \\
</json>}

\vspace{1em}
Now, please evaluate the **instance-centric compliance** of the provided image based on the above criteria and output the scores and reasoning in the specified **JSON** format.
\end{promptbox}

\end{document}